\documentclass[final]{cvpr}

\pdfoutput=1

\usepackage{times}
\usepackage{epsfig}
\usepackage{graphicx}
\usepackage{amsmath}
\usepackage{amssymb}

\DeclareMathOperator*{\argmin}{arg\,min}
\usepackage{adjustbox}
\usepackage{diagbox}
\usepackage{multirow}
\usepackage{caption}
\usepackage{subcaption}
\usepackage{booktabs}
\usepackage[table]{xcolor}
\usepackage{adjustbox}
\usepackage[pagebackref=true,breaklinks=true,colorlinks,bookmarks=false]{hyperref}

\def\cvprPaperID{1210} 
\def\confYear{CVPR 2021}
\begin{document}

\title{Semantic Segmentation with Generative Models: \\Semi-Supervised Learning and Strong Out-of-Domain Generalization}

\author{
Daiqing Li$^{1}\thanks{Correspondence to \{daiqingl,sfidler\}@nvidia.com}$
\and
Junlin Yang$^{1,3}$
\and
Karsten Kreis$^{1}$
\and
Antonio Torralba$^{4}$
\and
Sanja Fidler$^{1,2,5}$
\and \\[-2mm]
$^1$ NVIDIA \hspace{1.5mm}
$^2$ University of Toronto \hspace{1.5mm}
$^3$ Yale University \hspace{1.5mm}
$^4$ MIT \hspace{1.5mm}
$^5$ Vector Institute \hspace{1.5mm}
}

\maketitle

\begin{abstract}
Training deep networks with limited labeled data while achieving a strong generalization ability is key in the quest to reduce 
human annotation efforts.
This is the goal of semi-supervised learning, 
which exploits more widely available unlabeled data to complement small labeled data sets. 
In this paper, we propose a novel framework for discriminative pixel-level tasks using a generative model of both images and labels. Concretely, we learn a generative adversarial network that captures the joint image-label 
distribution and is trained efficiently using a large set of unlabeled images supplemented with only few labeled ones.
We build our architecture on top of StyleGAN2~\cite{karras2019analyzing}, 
augmented with a label synthesis branch.
Image labeling at test time is achieved by first embedding the target image into the joint latent space via an encoder network and test-time optimization,
and then generating the label from the inferred embedding.
We evaluate our approach in two important domains: medical image segmentation and part-based face segmentation. We demonstrate strong in-domain performance 
compared to several baselines, and are the first to showcase extreme out-of-domain generalization,
such as transferring from CT to MRI in medical imaging, and photographs of real faces to paintings, sculptures, and even cartoons and animal faces. Project Page: \url{https://nv-tlabs.github.io/semanticGAN/}


\end{abstract}

\vspace{-5mm}
\section{Introduction}
\vspace{-1mm}

Deep learning is now powering the majority of computer vision applications ranging from autonomous driving~\cite{neuralmotionplanner,liftsplat20} and medical imaging~\cite{ronneberger2015u,isensee2018nnunet} to image editing~\cite{park2019semantic,collins2020editing,zhu2020domain,viazovetskyi2020stylegan2,park2020swapping,plumerault2020controlling}. However, deep networks are extremely data hungry, typically requiring training on large-scale datasets to achieve high accuracy. Even when large datasets are available, generalizing the network's performance to out-of-distribution data, for example, on images captured by a different sensor, presents challenges, since deep networks tend to overfit to artificial statistics in the training data. Labeling large datasets, particularly for dense pixel-level tasks such as semantic segmentation, is already very time consuming. Re-doing the annotation effort each time the sensor changes is especially undesirable. This is particularly true in the medical domain, where pixel-level annotations are expensive to obtain (require highly-skilled experts), and where imaging sensors vary across sites. 
In this paper, we aim to significantly reduce the number of training data required for attaining successful performance, while achieving strong out-of-domain generalization. 

\begin{figure}[t!]
\vspace{-1.1mm}
    \begin{center}
        \includegraphics[width=0.98\linewidth]{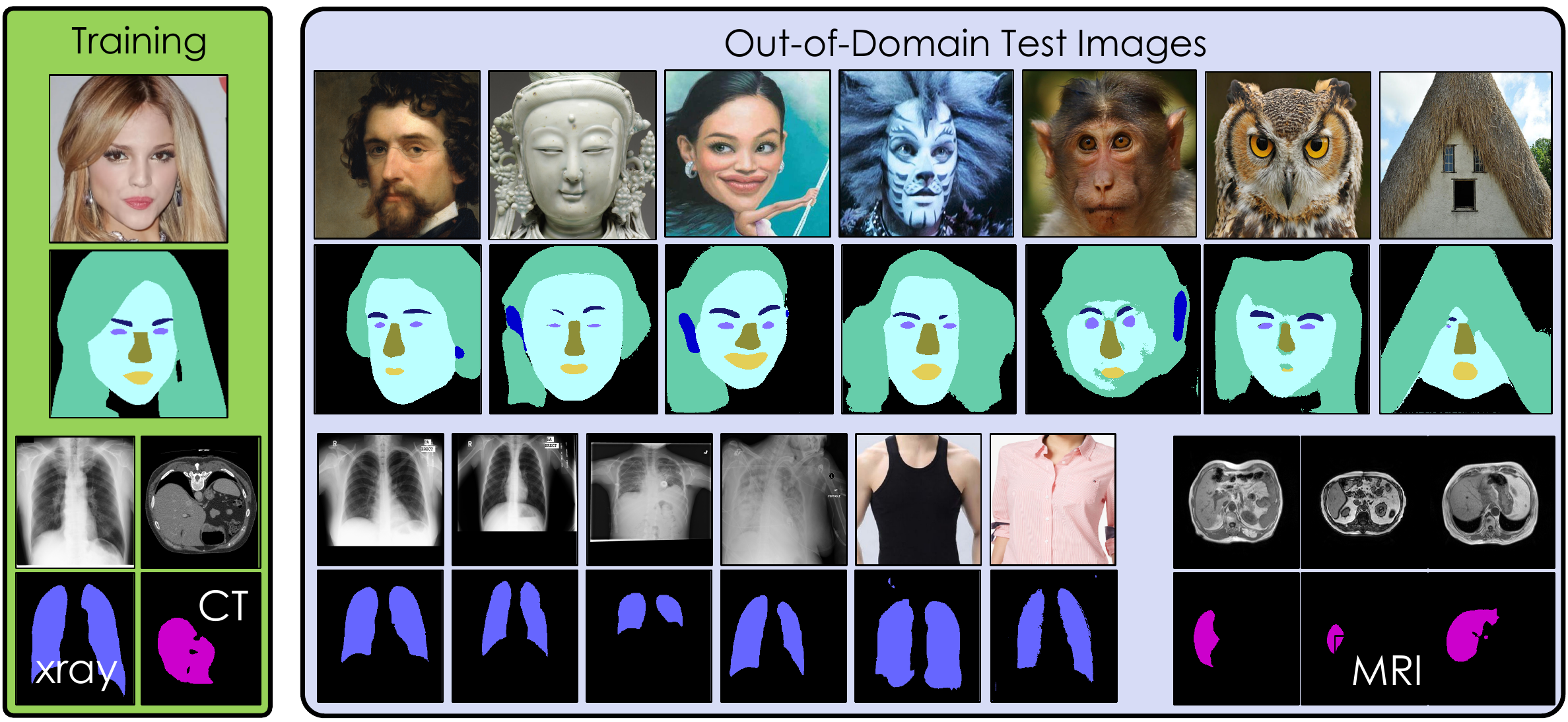}
    \end{center}
    \vspace{-6.4mm}
    \caption[]{
        \footnotesize \textbf{Out-of-domain Generalization.} Our model trained on real faces generalizes to paintings, sculptures, cartoons and even outputs plausible segmentations for animal faces. When trained on chest x-rays, it generalizes to multiple hospitals, and even  hallucinates lungs under clothed people.
        Our model also generalizes well from CT to MRI medical scans.
    }
    \label{fig:teaser}
    \vspace{-4.0mm}
\end{figure}

Semi-supervised learning (SSL) facilitates learning with small labeled data sets by augmenting the training set with large amounts of unlabeled data.
The literature on SSL is vast and some classical SSL techniques include pseudo-labeling~\cite{lee2013pseudo,bai2017semi,beyer2019s4l,sohn2020fixmatch}, consistency regularization~\cite{sajjadi2016reg,laine2017temporal,tarvainen2017mean,french2020milking,sohn2020fixmatch}, and various data augmentation techniques~\cite{berthelot2019mixmatch,Berthelot2020ReMixMatch,xie2020unsupervised} (also see Sec. \ref{sec:related}).
State-of-the-art SSL performance is currently achieved by contrastive learning, which aims to train powerful image feature extractors using unsupervised contrastive losses on image transformations~\cite{chen2020big,grill2020bootstrap,Misra_2020_CVPR,henaff2020dataefficient}. Once the feature extractors are trained, a smaller amount of labels is needed, since the features already implicitly encode semantic information. While SSL approaches have been more widely explored for classification, recent methods also tackle pixel-wise tasks~\cite{hung2018adversarial,mittal2019semi,ji2019invariant,ke2020guided,French2020,olsson2020classmix}. 

Although SSL techniques allow to train models with little labeled data, they usually do not explicitly model the distribution of the input data itself and therefore can still easily overfit to the training data, hampering their generalization capabilities. This is especially critical in semantic segmentation, where annotations are expensive and hence the available amount of labeled data can be particularly small.

To address this, we propose a fully generative approach based on a generative adversarial network (GAN) that models the \textit{joint} image-label distribution and synthesizes both images and their semantic segmentation masks. We build on top of the StyleGAN2~\cite{karras2019analyzing} architecture and augment it with a label generation branch. Our model is trained on a large unlabeled image collection and a small labeled subset using only adversarial objectives.
Test-time prediction is framed as first optimizing for the latent code that reconstructs the input image, and then synthesizing the label by applying the generator on the inferred embedding. 

We showcase our method in the medical domain and on human faces. It achieves competitive or better in-domain performance even when compared to heavily engineered state-of-the-art approaches, and shows significantly higher generalization ability on out-of-domain tests.
We also demonstrate the ability to generalize to domains that are drastically different from the training domain, such as going from CT to MRI volumes, and natural photographs of faces to sculptures, paintings and cartoons, and even animal faces (see Figure \ref{fig:teaser}). 

In summary, we make the following contributions: (i) We propose a novel generative model for semantic segmentation that builds on the state-of-the-art StyleGAN2 and naturally allows semi-supervised training. To the best of our knowledge, we are the first work that tackles semantic segmentation with a purely generative method that directly models the joint image-label distribution. (ii) We extensively validate our model in the medical domain and on face images. In the semi-supervised setting, we demonstrate results equal to or better than available competitive baselines. (iii) We show strong generalization capabilities and outperform our baselines on out-of-domain segmentation tasks by a large margin. (iv) We qualitatively demonstrate reasonable performance even on extreme out-of-domain examples.

\begin{figure*}[t!]
\begin{minipage}{0.65\linewidth}
    \begin{center}
        \includegraphics[width=0.92\linewidth]{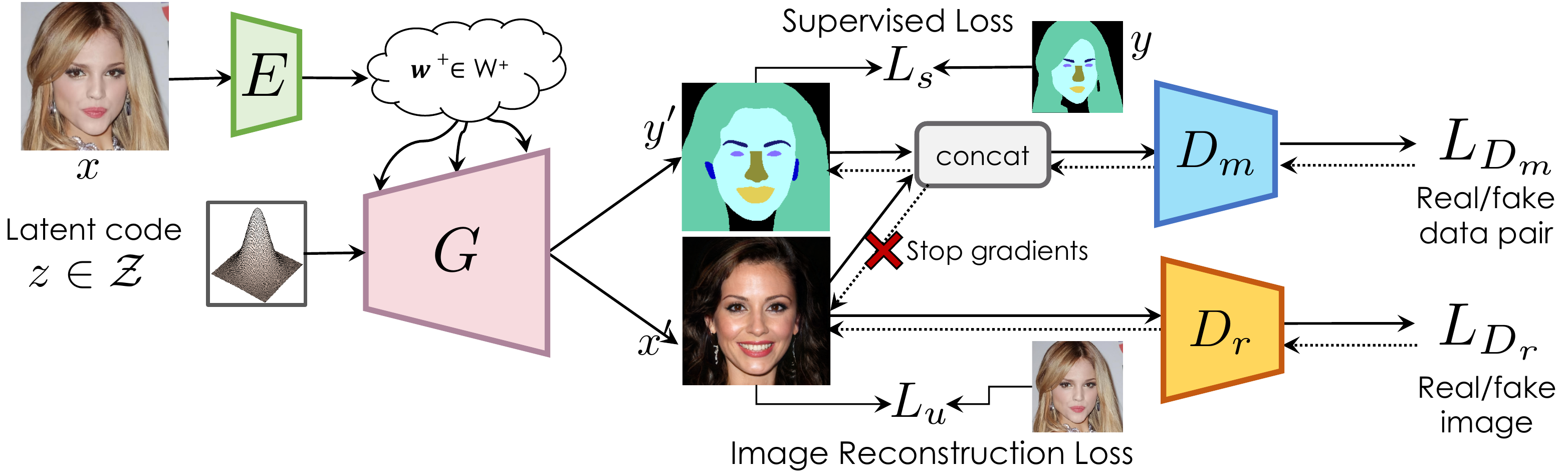}
    \end{center}
    \vspace{-5mm}
    \caption[]{
        \footnotesize \textbf{Model Overview.}  Generator $G$ and discriminators $D_m$ and $D_r$ are trained with adversarial objectives $\mathcal{L}_{G}$ (not indicated here), $\mathcal{L}_{D_m}$ and $\mathcal{L}_{D_r}$. We do not backpropagate gradients from ${D_m}$ into the generator's image synthesis branch. We train an additional encoder $E$ in a supervised fashion using image and mask reconstruction losses $\mathcal{L}_u$ and $\mathcal{L}_s$.
    }
    \label{fig:model}
    \end{minipage}
    \hfill
    \begin{minipage}{0.30\linewidth}
\begin{center}
   \includegraphics[width=0.99\linewidth]{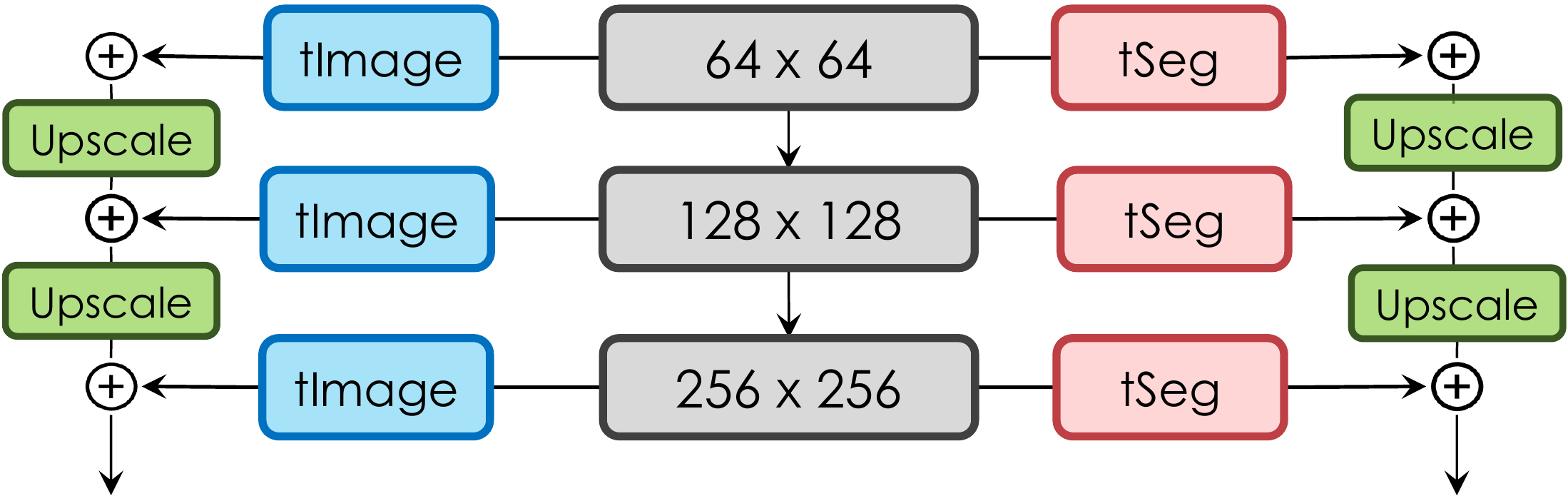}
\end{center}
\vspace{-4mm}
   \caption{\footnotesize \textbf{Generator Architecture.} We modify StyleGAN2's image synthesis network to also output masks. The \textit{tImage} and \textit{tSeg} blocks output intermediate images and segmentation masks at different resolutions, respectively. Both share the same style layers.}
\label{fig:generator}
\end{minipage}
    \vspace{-2.5mm}
\end{figure*}

\section{Related Work} \label{sec:related}
Our paper touches upon various topics, including medical image analysis, semantic segmentation, semi-supervised learning, generative modeling and neural network inversion.

\textbf{Semi-Supervised Learning and Semantic Segmentation}: in the medical domain, semi-supervised semantic segmentation has been tackled via pseudo-labeling~\cite{bai2017semi}, adversarial training~\cite{mondal2018fewshot,li2020transformation}, and transformation-consistency~\cite{li2020transformation} in a mean-teacher framework~\cite{tarvainen2017mean}.
In computer vision, \cite{luc2016semantic} is the first work using an adversarial objective to train a segmentation network. 
Later this idea was extended to semi-supervised setups via self-taught losses and discriminator feature matching~\cite{hung2018adversarial,mittal2019semi}. Recently, \cite{ke2020guided} proposed an approach using a flaw detector to approximate pixel-wise prediction confidence. 
Further relevant approaches to semi-supervised segmentation have been developed in weakly-supervised setups~\cite{Huang2018weakly,lee2019ficklenet,Zhang2019survey}.

For simpler classification tasks, a plethora of SSL methods have been developed, based on pseudo-labeling~\cite{lee2013pseudo,beyer2019s4l,pham2020meta}, self-supervision~\cite{beyer2019s4l},
entropy-minimization~\cite{Grandvalet2004semi}, consistency-regularization~\cite{sajjadi2016reg,laine2017temporal,tarvainen2017mean,french2020milking}, adversarial training~\cite{miyato2019vat}, data augmentation~\cite{xie2020unsupervised}, and combinations thereof~\cite{berthelot2019mixmatch,sohn2020fixmatch,Berthelot2020ReMixMatch}. However, current state-of-the-art semi-supervised methods are based on self-supervised learning with contrastive objectives~\cite{chen2020big,grill2020bootstrap,Misra_2020_CVPR,henaff2020dataefficient}. These approaches 
use unlabeled data in an often task-agnostic manner to learn general feature representations that can be ``finetuned'' using a smaller amount of labeled data. Related ideas have been applied to semi-supervised semantic segmentation~\cite{ji2019invariant} and tailored data augmentation strategies have been explored~\cite{French2020,olsson2020classmix}. Furthermore, many works employ carefully designed pretext tasks to learn useful representations from unlabeled images~\cite{doersch2015,Noroozi2016,gidaris2018unsupervised,goyal2019scaling}. Our method is related to these works in the sense that our task for learning strong features is image generation itself, instead of an auxiliary pretext task. 

The above works train discriminative models of the form $p(y|x)$, in contrast to our fully generative approach. However, generative approaches to SSL have been proposed before. \cite{kingma2014semi} leverages variational autoencoders and \cite{odena2016semisupervised,salimans2016gans} use GANs in which the discriminator distinguishes between different classes. 
A related approach to semi-supervised semantic segmentation uses generative models to augment the training data with additional synthesized data~\cite{souly2017semi,kar2019metasim,metasim20,fedsim}. Conceptually,~\cite{souly2017semi} trains a generator together with a pixel-wise discriminator network to perform segmentation, while~\cite{kar2019metasim,metasim20,fedsim} learn to generate synthetic 3D scenes by matching distributions of real and rendered imagery. In parallel work,~\cite{zhang21} exploit GANs to synthesize large labeled data datasets using very few labeled examples. 
In contrast, to the best of our knowledge, our method is the first fully generative approach to semantic segmentation that uses only adversarial objectives and no cross entropy terms and in which the generator models the \textit{joint} $p(x,y)$ and directly synthesizes images together with pixel-wise labels. We further use the generative model as a decoder of semantic outputs at test time, which we show leads to better generalization than prior and parallel work.

\textbf{Generator Inversion}: A critical part of our method is the effective inversion of the GAN generator at test-time to infer the latent embedding of a new image to be labeled. We are building on previous works that have studied this task before. Optmization-based methods iteratively optimize a reconstruction objective~\cite{zhu2016generative,Yeh2017,lipton2017precise,abdal2019image2stylegan,huh2020transforming,creswell2019inverting,Raj2019gan,plumerault2020controlling} or perform Markov chain Monte Carlo~\cite{fang2019co}, while encoder-based techniques directly map target images into the embedding space~\cite{perarnau2016invertible,donahue2016adversarial,Brock2017neural,dumoulin2017adversarially,richardson2020encoding}. Hybrid methods combine these ideas and initialize iterative optimization from an encoder prediction~\cite{zhu2016generative,bau2018gan,bau2019semantic,bau2019seeing,zhu2020domain}. These works primarily focus on image reconstruction and editing, while we use inferred embeddings for pixel-wise image labeling.


\textbf{Generative Models for Image Understanding}:
Our approach is in line with various works that explore the use of generative modeling in different forms for discriminative and image recognition tasks, an idea that dates back until at least \cite{ng2001} and has also been studied using early energy-based models~\cite{ranzato2011,Hinton2007ToRS,Srivastava2012}. \cite{ranzato2011,Hinton2007ToRS} train deep belief networks to model shapes and learn representations of images in the model's latent variables. These representations can then be used for image recognition. In~\cite{amodalVAE20}, a VAE was used for amodal object instance segmentation. These ideas are closely related to our method, which learns features in a GAN generator that can be used for semantic segmentation. 

Recently, \cite{Yeh2017,fang2019co,gu2020mganprior} demonstrated impressive inpainting as well as colorization and super-resolution results using GANs. In fact, also our method can be interpreted as ``inpainting'' of missing labels using a generative model of the joint image-label distribution in a similar manner. 
Along a different line of research, \cite{Grathwohl2020Your,liu2020hybrid,huang2020neural} found that generative training of deep classification networks results in better calibrated and more robust models, which is consistent with the strong generalization capabilities we observe in our method.

These related works motivate to also treat semantic segmentation as a generative modeling problem.

\vspace{-2.5mm}
\section{Method}
We first provide a conceptual overview over our method and discuss its motivation and advantages. Then, we explain the model architecture, training, and inference in detail.

\vspace{-1mm}
\subsection{Overview}
Traditional neural network-based semantic segmentation methods~\cite{chen2017rethinking} learn a function $f: \mathcal{X}\rightarrow \mathcal{Y}$, mapping images $x\in \mathcal{X}$ to pixel-wise target labels $y\in \mathcal{Y}$. The goal of learning is to maximize the conditional probability $p(y|x)$. This requires large labeled data sets and is prone to overfitting when training with limited amounts of annotated images.

We propose to instead model the \textit{joint} distribution of images and labels $p(x,y)$ with a GAN-based generative model. In the GAN framework, $p(x,y)$ is implicitly defined as the distribution obtained when mapping latent variables $z$ drawn from a noise distribution $p(z)$ through a deterministic generator $G(z):\mathcal{Z}\rightarrow(\mathcal{X},\mathcal{Y})$ that outputs both images $x$ and labels $y$.
In this setup, a latent vector $z$ explains both the image and its labels, and, given $z$, image and labels are conditionally independent. Hence, we can label a new image $x^*$ by first inferring its embedding $z^*$ via an auxiliary encoder and test-time optimization, and then synthesize the corresponding pixel-wise labels $y^*$ (in practice, we are working directly in StyleGAN2's $\mathcal{W}^+$-space instead of the ``Normal'' $\mathcal{Z}$-space). See Figure \ref{fig:model} for an overview.

\vspace{-1.2mm} \label{sec:methodmotivation}
\subsection{Motivation}
Our fully generative approach to semantic segmentation has several advantages over traditional methods that directly model the conditional $p(y|x)$.

\textbf{Semi-supervised Training}: Intuitively, a model that can generate realistic images should know how to generate the corresponding pixel-wise labels as well, as they are just capturing semantic information already present in the image itself. This is analogous to rendering, where, if we know how to render a given scene, generating labels of interest, such as segmentation or depth, is simple. 
A GAN can be viewed as a neural renderer, where the embeddings $z$ completely encode and describe the images to be synthesized via a neural network~\cite{style3d}. 
This connection suggests a similar strategy: If we know how to generate images, the GAN should be able to easily generate associated labels as well. This implies that the feature representations learnt by the GAN 
can be expected to be useful also for pixel-wise labeling tasks. Hence, we can simply augment the generator with a small additional branch that synthesizes labels from the same features used for image generation. 
A major benefit of this approach is that training the GAN itself only requires images without labels. A small amount of labels is only necessary for training the small labeling function on top of the main GAN architecture.
Therefore, this setup naturally allows for efficient semi-supervised training. Furthermore, by jointly training the GAN for image and label synthesis, its features can be further ``finetuned'' for semantic label generation.

Note that we can also view this setup as parameter sharing: Given an embedding $z$, the label-generating function shares nearly all its parameters with the image-generating function and only adds few additional parameters that are solely trained with labeled data.

\textbf{Generalization}: After training, we expect the model to synthesize plausible image-label pairs for all embeddings $z$ within the noise distribution $p(z)$, from which we drew samples during training. Therefore, we will likely be able to successfully label any new images, whose embeddings are in $p(z)$ or sufficiently close. Furthermore, the GAN never sees the same input repeatedly during training, as its input is the resampled noise $z$. Hence, it learns a smooth generator function over the complete latent distribution $p(z)$. In contrast, a purely conditional model $p(y|x)$ is much more likely to overfit to the limited labeled training data and does not take into account the distribution $p(x)$ of the data itself. For these reasons, our generative approach can be expected to show significantly better generalization capabilities beyond the training data and even beyond the training domain, which we validate in our experiments.

\vspace{-1.5mm}
\subsection{Model} \label{sec:model}
We build our model on top of StyleGAN2~\cite{karras2019analyzing}, the current state-of-the-art GAN for image synthesis. It is based on its successor StyleGAN~\cite{karras2019style} and proposes several modifications, such as latent space path-length regularization to encourage generator smoothness and a redesign of instance normalization to remove generation artifacts. Furthermore, the previous progressive growing strategy~\cite{karras2017progressive} is abandoned in favor of a residual skip-connection design. The model achieves remarkable image synthesis quality and has found important applications for example in image editing~\cite{viazovetskyi2020stylegan2}. 
We now explain our model design in detail.

\textbf{Generator:} Our generator is based on StyleGAN2's generator
with residual skip-connection design~\cite{karras2019analyzing}. We add an additional branch at each style layer to output a segmentation mask $y$ along with the image output $x$ (Figure \ref{fig:generator}). Like standard StyleGAN2, our generator takes random noise vectors $z\in \mathcal{Z}$ following a simple Normal distribution $p(z)=\mathcal{N}(0,I)$ as input and first transforms them via a fully-connected network to a more complex distribution $p(w)$ in a space usually denoted as $\mathcal{W}$~\cite{karras2019style}. After an affine transformation, these complex noise variables are then fed to the generator's main style layers, which output images $x\in\mathcal{X}$ and pixel-wise labels $y\in \mathcal{Y}$. We can formally define this as $G: \mathcal{Z} \rightarrow \mathcal{W} \rightarrow (\mathcal{X},\mathcal{Y})$.

\textbf{Discriminators:} We have two discriminators $D_r$ and $D_m$. Specifically, $D_r: \mathcal{X} \rightarrow \mathbb{R}$ is applied on real and generated images, encouraging the generator to produce realistic images. It follows the residual architecture of \cite{karras2017progressive,karras2019style}. $D_m: (\mathcal{X}, \mathcal{Y}) \rightarrow \mathbb{R}$ consumes both images and pixel-wise label masks via concatenation and discriminates between generated and real image-label pairs. This enforces alignment between synthesized images and labels, as non-aligned image-label pairs could be easily detected as ``fake''. To enforce strong consistency between images and labels, we are using the multi-scale patch-based discriminator architecture from \cite{wang2018high} for $D_m$.

\textbf{Encoder and $\mathcal{W}^+$-space:} During inference, we first need to infer a new image's embedding. Instead of performing inference in $\mathcal{Z}$-space, it has been shown that it is beneficial to instead directly work in $\mathcal{W}$-space and to model all noise vectors $w$ \textit{independently}, unlike in training, where the \textit{same} $w$ is provided to all style layers~\cite{abdal2019image2stylegan}. When modeling the $w$'s independently for each style layer, we can interpret this as an extended space, which is usually denoted as $\mathcal{W}^+$ with elements $w^+$. We are following this previous work and perform embedding inference in $\mathcal{W}^+$. Below, when writing $G(w^+)$, we indicate generation directly based on $w^+$, instead of samples $z\in \mathcal{Z}$.

As explained below, we infer an image's $w^+$ embedding via test-time optimization. To speed up this optimization process and provide a strong initialization, we are using an additional encoder $E: \mathcal{X} \rightarrow \mathcal{W}^+$, mapping images $x$ directly to $\mathcal{W}^+$-space. Its architecture is based on \cite{richardson2020encoding}, which uses a feature pyramid network~\cite{lin2017feature} as backbone to extract multi-level features. A small fully convolutional network is used to map those features to $\mathcal{W}^+$-space (see Figure \ref{fig:model}).

\vspace{-1.7mm}
\subsection{Training}

We utilize a large unlabeled data set $D_{u} = \{x_1,...,x_n\}$ and a small labeled data set $D_{l} = \{(x_1,y_1),...,(x_k,y_k)\}$, with $k \ll n$. We are training in two stages and train generator and discriminators first and encoder second.

\textbf{Loss Function:} The generator and the discriminators are trained with the following standard GAN objectives:
\begin{equation}
    \begin{split}
        \mathcal{L}_{D_r} = & \mathop{\mathbb{E}}_{x_r \sim D_{u}}[\log D_r(x_r)] \\ & + \mathop{\mathbb{E}}_{(x_f,\cdot) = G(z), z\sim p(z)}[\log (1 - D_r(x_f))]
    \end{split}
\end{equation}
\begin{equation}
    \begin{split}
    \mathcal{L}_{D_m} = & \mathop{\mathbb{E}}_{(x_r,y_r) \sim D_{l}}[\log D_m(x_r,y_r)] \\ & + \mathop{\mathbb{E}}_{(x_f,y_f) = G(z), z\sim p(z)}[\log (1 - D_m(x_f,y_f))]
    \end{split}
\end{equation}
\begin{equation} \label{eq:gen_obj}
    \begin{split}
    \mathcal{L}_{G} = & \mathop{\mathbb{E}}_{(x_f,\cdot) = G(z), z\sim p(z)}[\log (1 - D_r(x_f))] \\ & + \mathop{\mathbb{E}}_{(x_f,y_f) = G(z), z\sim p(z)}[\log (1 - D_m(x_f,y_f))]
    \end{split}
\end{equation}



The objective of the discriminators $D_r$ and $D_m$ is to maximize $\mathcal{L}_{D_r}$ and $\mathcal{L}_{D_m}$ respectively, while the objective of the generator $G$ is to minimize $\mathcal{L}_G$. The second term in Eq.~(\ref{eq:gen_obj}) leads to gradients in both the image and label branch of the generator. These gradients are produced by $D_m$ and encourage adjustment of synthesized labels and images. However, we want the synthesized label to be adjusted to match the synthesized image, instead of the other way, \ie perturbing image generation to match the labels. Therefore, we are stopping gradient backpropagation into the generator via the image synthesis branch from the second term in Eq.~(\ref{eq:gen_obj}). In this way, the image generation branch is trained purely via the image generation task with feedback from $D_r$ (first term in Eq.~(\ref{eq:gen_obj})), using the complete data set including the unlabeled images. At the same time, the GAN's main features in the style layers, to which both the image and label synthesis branches are connected, are still experiencing feedback from both the image and the label synthesis branch. Due to this joint training strategy, the generator learns feature representations useful both for realistic image synthesis and corresponding label generation. 
Note that we use only adversarial losses. There are no pair-wise losses for segmentation, such as cross-entropy between pairs of real and generated label masks, at all.

\textbf{Encoder:} 
When training the encoder $E: \mathcal{X} \rightarrow \mathcal{W}^+$, we freeze the generator $G$. The encoder training objective is
\begin{equation}
    \mathcal{L}_E = \mathcal{L}_s +  \mathcal{L}_u,
\end{equation}
$\mathcal{L}_s$ is the supervised loss on labeled images, defined as:
\begin{equation}
    \mathcal{L}_s = 
    \mathop{\mathbb{E}}_{(x, y) \sim D_l} \mathbf{H} (y, G_y(E(x))) + \mathbf{DC} (y, G_y(E(x)))
\end{equation}
with $\mathbf{H}(\cdot,\cdot)$ denoting a pixel-wise cross-entropy loss summed over all pixels and $\mathbf{DC}(\cdot,\cdot)$ the dice loss as in \cite{isensee2018nnunet}. The unsupervised loss $\mathcal{L}_u$ is
\begin{equation}
\vspace{-1mm}
    \begin{split}
        \mathcal{L}_u = & 
        \mathop{\mathbb{E}}_{x \sim D_l \cup D_u} \mathcal{L}_{\textrm{LPIPS}}(x, G_x(E(x))) \\ & + \lambda_1 ||x - G_x(E(x))||_2^2
    \end{split}
\end{equation}
with $\lambda_1$ a hyperparameter trading off  different loss contributions, $G_x$ denoting the generator's image backbone and $G_y$ the label generation branch. $\mathcal{L}_{\textrm{LPIPS}}(x_1,x_2)$ is the \textit{Learned Perceptual Image Patch Similarity} (LPIPS) distance~\cite{zhang2018unreasonable}, which measures L2 distance in the feature space of an ImageNet-pretrained VGG19 network.
With the above objective, we are training the encoder to map images $x$ to embeddings $w^+\in \mathcal{W}^+$, which re-generate the input images and, for labeled data, also the pixel-wise label masks. 

\vspace{-1mm}
\subsection{Inference} \label{sec:inference}
At inference time, we are given a target image $x^*$ and our goal is to find the optimal pixel-wise labels $y^*$. As explained above, we first embed the target image into the generator's embedding space, for which we choose $\mathcal{W}^+$ instead of $\mathcal{Z}$. To this end, we are mapping the image $x^*$ to $\mathcal{W}^+$ using the encoder $E$ and then solve the inversion objective
\begin{equation} \label{eq:testtime_opt}
\begin{split}
    w^{+*} = \argmin_{w^+ \in \mathcal{W}^+}[& \mathcal{L}_{\textrm{reconst}}(x^*,G_x(w^+))\\ & + \lambda_2 ||w^+ - E(G(w^+))||_2^2 ]
\end{split}
\end{equation}
iteratively via gradient descent-based methods. The first term in Eq. (\ref{eq:testtime_opt}) optimizes for reconstruction quality of the given image and the second term regularizes the optimization trajectory to stay in the training domain, where the encoder was training to approximately invert the generator. This strategy was recently proposed in \cite{zhu2020domain}.
This regularization, controlled by the hyperparameter $\lambda_2$, can be particularly beneficial when performing labeling of images outside the training domain. In this case, purely optimizing for reconstruction quality can result in $w^{+*}$ values that lie far outside the distribution $p(w^+)$ of embeddings $w^+$ encountered during training. Since the labeling branch is not trained for such $w^{+*}$, the predicted label may be incorrect. 
One may suggest to instead directly regularize with $p(w^+)$, however, this is not easily possible, since $p(w^+)$ is not actually a tractable distribution. It is only implicitly defined by mapping samples from $p(z)$ though the fully-connected noise transformation layers of StyleGAN2.

For the reconstruction term $\mathcal{L}_{\textrm{reconst}}$ we follow \cite{karras2019analyzing} and use LPIPS together with a per-pixel L2 term:
\begin{equation} \label{eq:opt_rec_loss}
    \mathcal{L}_{\textrm{reconst}}(x,x^*) = \mathcal{L}_{\textrm{LPIPS}}(x,x^*) + \lambda_3 ||x-x^*||^2_2
\end{equation}
where $\lambda_3$ is another hyperparameter.
After obtaining $w^{+*}$, we pass it back to the generator to get $G(w^{+*})=(x^{inv}, y^{inv})$. Since $w^{+*}$ was optimized to minimize the reconstruction error between $x^{inv}$ and $x^*$, we have $x^{inv}\approx x^*$. Furthermore, as the generator was trained to align synthesized segmentation labels and images, we can expect $y^{inv}$ to be a correct label of the reconstructed image $x^{inv}$. Hence, the generated segmentation mask $y^{inv}$ is the almost optimal segmentation $y^* \approx y^{inv}$ of the target image $x^*$.

Note that we can also look at our inference protocol from a fully probabilistic perspective, where we find the maximum of the log posterior distribution over embeddings given an image. In the supplemental material, we discuss this in more detail and how it relates to other works.

\definecolor{orange}{rgb}{1,0.8,0.8}
\begin{figure}[t!]
\begin{adjustbox}{width=\linewidth,center}
\scriptsize
\addtolength{\tabcolsep}{-4pt}
\begin{tabular}{cccccccc}
\multicolumn{2}{c}{Faces} & 
\multicolumn{2}{c}{Chest X-ray} &
\multicolumn{2}{c}{Skin Lesion } &
\multicolumn{2}{c}{Liver CT} 
\\
\includegraphics[width=0.125\linewidth]{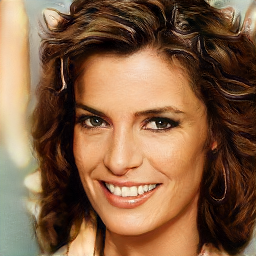}
&
\includegraphics[width=0.125\linewidth]{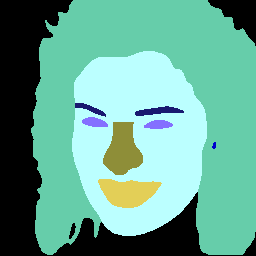}
&
\includegraphics[width=0.125\linewidth]{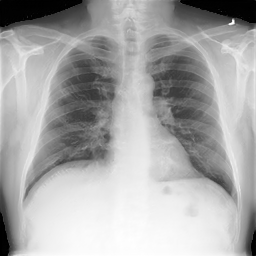}
&
\includegraphics[width=0.125\linewidth]{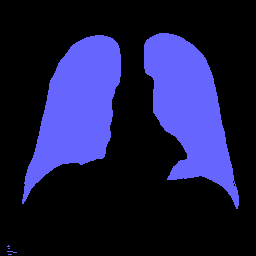}
&
\includegraphics[width=0.125\linewidth]{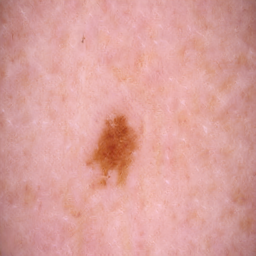}
&
\includegraphics[width=0.125\linewidth]{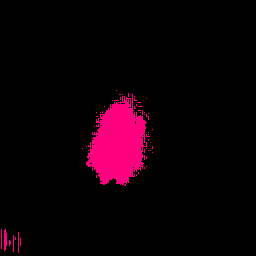}
&
\includegraphics[width=0.125\linewidth]{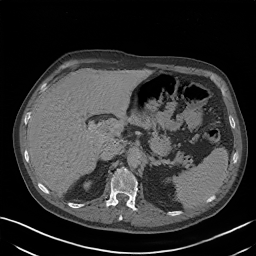}
&
\includegraphics[width=0.125\linewidth]{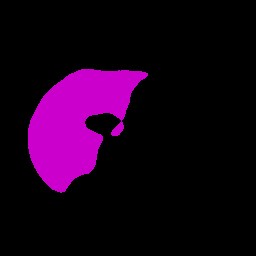}
\\

\includegraphics[width=0.125\linewidth]{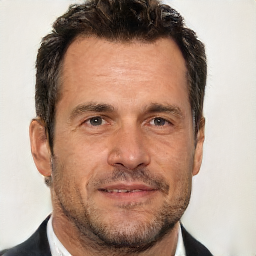}
&
\includegraphics[width=0.125\linewidth]{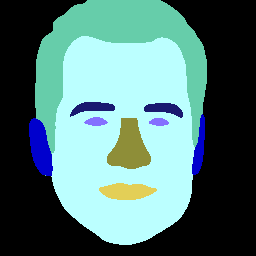}
&
\includegraphics[width=0.125\linewidth]{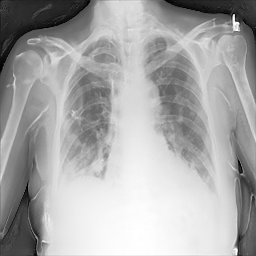}
&
\includegraphics[width=0.125\linewidth]{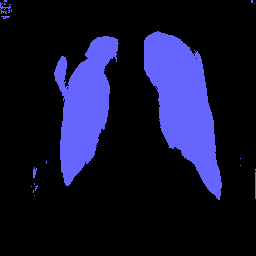}
&
\includegraphics[width=0.125\linewidth]{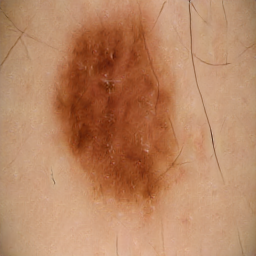}
&
\includegraphics[width=0.125\linewidth]{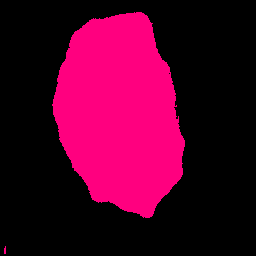}
&
\includegraphics[width=0.125\linewidth]{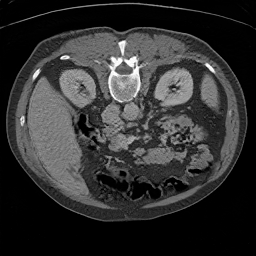}
&
\includegraphics[width=0.125\linewidth]{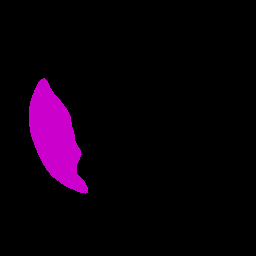}
\\

\includegraphics[width=0.125\linewidth]{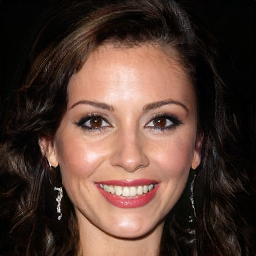}
&
\includegraphics[width=0.125\linewidth]{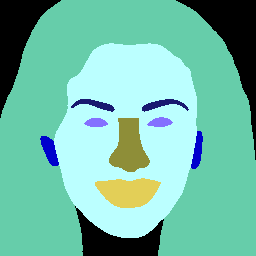}
&
\includegraphics[width=0.125\linewidth]{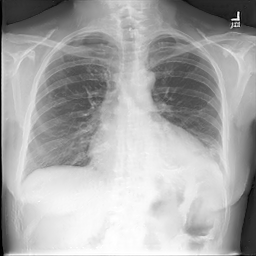}
&
\includegraphics[width=0.125\linewidth]{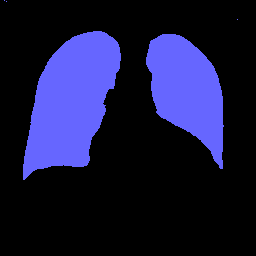}
&
\includegraphics[width=0.125\linewidth]{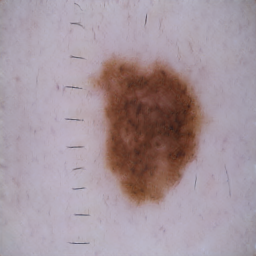}
&
\includegraphics[width=0.125\linewidth]{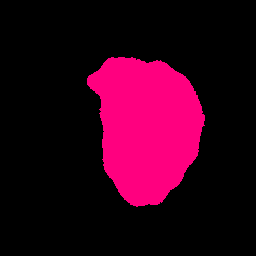}
&
\includegraphics[width=0.125\linewidth]{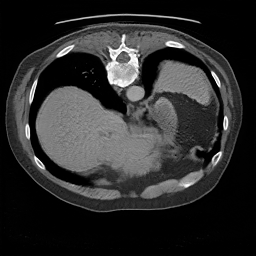}
&
\includegraphics[width=0.125\linewidth]{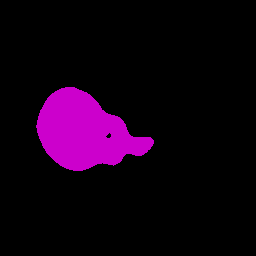}
\\

\end{tabular}
\end{adjustbox}
\vspace{-4mm}
\caption{\footnotesize {\bf Synthetic Samples} of image and pixel-wise segmentation label pairs from our generator for multiple datasets.}
\label{fig:qual_syn_samples}
\vspace{-3mm}
\end{figure}

\vspace{-3mm}
\section{Experiments}
Our approach is limited by the expressivity of the generative model. Although GANs have achieved outstanding synthesis quality for ``unimodal'' data such as images of faces, current generative models cannot model highly complex data, such as images of vivid outdoor scenes. Hence, our method is not applicable to such data. Therefore, in our experiments we focus on human faces as well as the medical domain, where most images can be successfully modeled by StyleGAN2 (see Figure \ref{fig:qual_syn_samples}), and where annotation is particularly expensive, as it relies on highly skilled experts.

\begin{table*}[t!]
\vspace{-4mm}
\begin{minipage}{0.7\linewidth}
\begin{center}
\resizebox{1\linewidth}{!}{
\rowcolors{2}{white}{gray!15}
\begin{tabular}{lcccccccccccccc}
\toprule
\rowcolor{white}
 & 
\multicolumn{4}{c}{Trained with \textbf{9} labeled data samples} && 
\multicolumn{4}{c}{Trained with \textbf{35} labeled data samples} &&
\multicolumn{4}{c}{Trained with \textbf{175} labeled data samples} \\
\cmidrule{2-5} \cmidrule{7-10} \cmidrule{12-15} 
\rowcolor{white}
Method& JSRT & NLM & NIH & SZ && JSRT & NLM & NIH & SZ && JSRT & NLM & NIH & SZ \\ 
\midrule
U-Net            & 0.9318 & 0.8605 & 0.6801 & 0.9051 
                && 0.9308 & 0.8591 & 0.7363 & 0.8486 
                && 0.9464 & 0.9143 & 0.7553 & 0.9005\\
                
                
DeepLab         & 0.9006 & 0.6324 & 0.7361 & 0.8124 
                && 0.9556 & 0.8323 & 0.8099 & 0.9138 
                && 0.9666 & 0.8175 & 0.8093 & 0.9312\\
                
\midrule
MT                                   & 0.9239 & 0.8287 & 0.7280 & 0.8847 
                                     && 0.9436 & 0.8239 & 0.7305 & 0.8306 
                                     && 0.9604 & 0.8626 & 0.7893 & 0.8846 \\
                                     
AdvSSL                               & 0.9328 & 0.8500 & 0.7720 & 0.8901 
                                     && 0.9552 & 0.8191 & 0.5298 & 0.8968 
                                     && \textbf{0.9684} & 0.8344 & 0.7627 & 0.8846 \\
                                     
GCT                                  & 0.9235 & 0.6804 & 0.6731 & 0.8665 
                                     && 0.9502 & 0.8327 & 0.7527 & 0.9184 
                                     && 0.9644 & 0.8683 & 0.7981 & 0.9393 \\

\midrule

Ours-NO  & 0.9464 & 0.9303 & 0.9097 & 0.9334 && 0.9471 & 0.9294 & 0.9223 & 0.9409 && 0.9465 & 0.9232 & 0.9204 & 0.9403 \\
Ours     & \textbf{0.9591} & \textbf{0.9464} & \textbf{0.9133} & \textbf{0.9362} && \textbf{0.9668} & \textbf{0.9606} & \textbf{0.9322} & \textbf{0.9485} && 0.9669 & \textbf{0.9509} & \textbf{0.9294} & \textbf{0.9469} \\
\bottomrule

\end{tabular}
}
\end{center}
\end{minipage}
\hspace{1mm}
\begin{minipage}{0.282\linewidth}
\vspace{3mm}
\caption{\footnotesize \textbf{Chest X-ray Lung Segmentation.} Numbers are DICE scores. CXR14 \cite{wang2017chestx} JSRT \cite{shiraishi2000development}are the in-domain data set, on which we both train and evaluate. We also evaluate on additional out-of-domain datasets (NLM \cite{jaeger2014two}, NIH \cite{stirenko2018chest}, SZ \cite{jaeger2014two}, details in supplemental material). Ours as well as the semi-supervised methods use additional 108k unlabeled data samples.}
\label{tab:xray-out-domain}
\end{minipage}
\vspace{-1mm}
\end{table*}
\begin{table*}[t!]
\vspace{-8mm}
\begin{minipage}{0.7\linewidth}
\begin{center}
\resizebox{1\linewidth}{!}{
\rowcolors{2}{white}{gray!15}
\begin{tabular}{lcccccccccccccc}
\toprule
\rowcolor{white}
 & 
\multicolumn{4}{c}{Trained with \textbf{40} labeled data samples} & & 
\multicolumn{4}{c}{Trained with \textbf{200} labeled data samples} & &
\multicolumn{4}{c}{Trained with \textbf{2000} labeled data samples} \\
\cmidrule{2-5} \cmidrule{7-10} \cmidrule{12-15}
\rowcolor{white}
Method& ISIC & PH2 & IS & Quest && ISIC & PH2 & IS & Quest & & ISIC & PH2 & IS & Quest \\ 
\midrule
U-Net     & 0.4935 & 0.4973 & 0.3321 & 0.0921 &
         & 0.6041 & 0.7082 & 0.4922 & 0.1916 &
         & 0.6469 & 0.6761 & 0.5497 & 0.3278\\
DeepLab  & 0.5846 & 0.6794 & 0.5136 & 0.1816 &
         & 0.6962 & 0.7617 & 0.6565 & 0.4664 &
         & 0.7845 & 0.8080 & 0.7222 & 0.6457\\
\midrule
MT                                   & 0.5200 & 0.5813 & 0.4283 & 0.1307 &
                                     & 0.7052 & 0.7922 & 0.6330 & 0.4149 & 
                                     & 0.7741 & 0.8156 & 0.6611 & 0.5816 \\
AdvSSL                               & 0.5016 & 0.5275 & 0.5575 & 0.1741 &
                                     & 0.6657 & 0.7492 & 0.6087 & 0.3281 & 
                                     & 0.7388 & 0.7351 & 0.6821 & 0.6178 \\
GCT                                  & 0.4759 & 0.4781 & 0.5436 & 0.1611 &
                                     & 0.6814 & 0.7536 & 0.6586 & 0.3109 & 
                                     & 0.7887 & 0.8248 & 0.7104 & 0.5681 \\
                                     
\midrule
Ours-NO  & 0.6987 & 0.7565 & 0.7083 & 0.5060 &
         & 0.7517 & \textbf{0.8160} & 0.7150 & 0.6493 &
         & 0.7855 & 0.8087 & 0.6876 & 0.6350 \\
Ours     & \textbf{0.7144} & \textbf{0.7950} & \textbf{0.7350} & \textbf{0.5658} &
         & \textbf{0.7555} & 0.8154 & \textbf{0.7388} & \textbf{0.6958} &
         & \textbf{0.7890} & \textbf{0.8329} & \textbf{0.7436} & \textbf{0.6819} \\
\bottomrule

\end{tabular}
}
\end{center}
\end{minipage}
\hspace{1mm}
\begin{minipage}{0.282\linewidth}
\vspace{3mm}
\caption{\footnotesize \textbf{Skin Lesion Segmentation.} Numbers are JC index. Here, ISIC \cite{codella2019skin} is the in-domain data set, on which we train and also evaluate. Additionally, we perform segmentation on three out-of-domain datasets (PH2 \cite{mendoncca2013ph}, IS \cite{glaister2013automatic}, Quest \cite{glaister2013automatic}, details in supplemental material). Ours as well as the semi-supervised methods use additional $\approx$33k unlabeled data samples.}
\label{tab:isic-out-domain}
\end{minipage}
\vspace{-5mm}
\end{table*}
\begin{table*}[t!]
\vspace{-2.5mm}
\vspace{-1mm}
\begin{minipage}{0.7\linewidth}
\begin{adjustbox}{width=\textwidth,center}
\rowcolors{2}{white}{gray!15}
\addtolength{\tabcolsep}{-4pt}
\begin{tabular}{lccccccccccc}
\toprule
 & 
\multicolumn{3}{c}{Trained with \textbf{8} labeled examples} & &
\multicolumn{3}{c}{Trained with \textbf{20} labeled examples} & &
\multicolumn{3}{c}{Trained with \textbf{118} labeled examples} \\
\cmidrule{2-4} \cmidrule{6-8} \cmidrule{10-12}
Method& CT & MRI T1-in & MRI T1-out && CT & MRI T1-in & MRI T1-out && CT & MRI T1-in & MRI T1-out \\ 
\midrule
U-Net  & 0.7610 & 0.2568 & 0.3293 && 0.8229 & 0.3428 & 0.2310 && 0.8680 & 0.4453 & 0.4177 \\
\midrule
Ours-NO  & 0.8036 & 0.4811 & 0.5135 && 0.8462 & \textbf{0.5538} & 0.4511 && 0.8603 & 0.5055 & \textbf{0.5633} \\
Ours     & \textbf{0.8747} & \textbf{0.5565} & \textbf{0.5678} && \textbf{0.8961} & 0.4989 & \textbf{0.4575} && \textbf{0.9169} & \textbf{0.5097} & 0.5243 \\
\bottomrule

\end{tabular}
\end{adjustbox}
\end{minipage}
\hspace{2mm}
\begin{minipage}{0.282\linewidth}
\vspace{2mm}
\caption{\footnotesize {\bf CT-MRI Transfer Liver Segmentation.} Numbers are DICE per patient. Here, CT is the in-domain data set. We evaluate on unseen MRI data~\cite{kavur2020chaos} for liver segmentation task. Ours uses additional 70 volumes from LITS2017~\cite{bilic2019liver} testing set as unlabeled data samples.}
\label{tab:liver-out-domain}
\end{minipage}
\vspace{-6mm}
\end{table*}

We test our method on three medical tasks, chest X-ray segmentation, skin lesion segmentation, and cross-domain computer tomography to magnetic resonance image (CT-MRI) liver segmentation, as well as face part segmentation. For each task, we assume we have access to a small labeled and a relatively large unlabeled data set. We test our model on in-domain and several out-of-domain data sets. In the following, we explain our experimental setup, and report results by qualitatively and quantitatively comparing to strong baselines. We also analyze the value of labeled, unlabeled and synthetic data. 
Implementation details are in Appendix.

\vspace{-1mm}
\subsection{Setup}
\textbf{Datasets.} 
For chest x-ray segmentation, we use two in-domain datasets (for labeled and unlabeled data), on which we train the model. We evaluate on three additional out-of-domain datasets. The datasets vary in terms of sensor quality and patient poses. We follow a similar approach for skin lesion segmentation and combine two datasets for training and evaluate additionally on three out-of-domain datasets. For the cross-domain CT-MRI liver segmentation task we use a CT dataset as our in-domain training data and also evaluate on two MRI datasets.
For face part segmentation, we use the CelebA dataset~\cite{liu2018large}. Furthermore, for out-of-domain evaluation we randomly selected 40 images from the MetFaces dataset~\cite{karras2020training}, a collection of human face paintings and sculptures, and manually annotated them following the labeling protocol for CelebA. 

\textbf{Metrics.}
For chest X-ray and CT-MRI liver 
segmentation, we report per-patient DICE scores, the default metric used in the literature for this task. For skin lesion segmentation, we report the per-patient JC index, following the ISIC challenge~\cite{rotemberg2020patientcentric}. For face part segmentation, we use mean Intersection over Union (mIoU) over all classes, excluding the background class. mIoU is the most widely used metric in computer vision for segmentation tasks. 

\begin{figure*}[t!]
\vspace{-2mm}
\begin{minipage}{0.47\linewidth}
\begin{adjustbox}{center}
\footnotesize
\addtolength{\tabcolsep}{-4pt}
\begin{tabular}{cccccccc}
& image & GT  & DeepLab& MT&AdvSSL& GCT &Ours \\
%
\rotatebox{90}{\scriptsize \hspace{-1mm}In-Domain}

&
\includegraphics[width=0.125\linewidth]{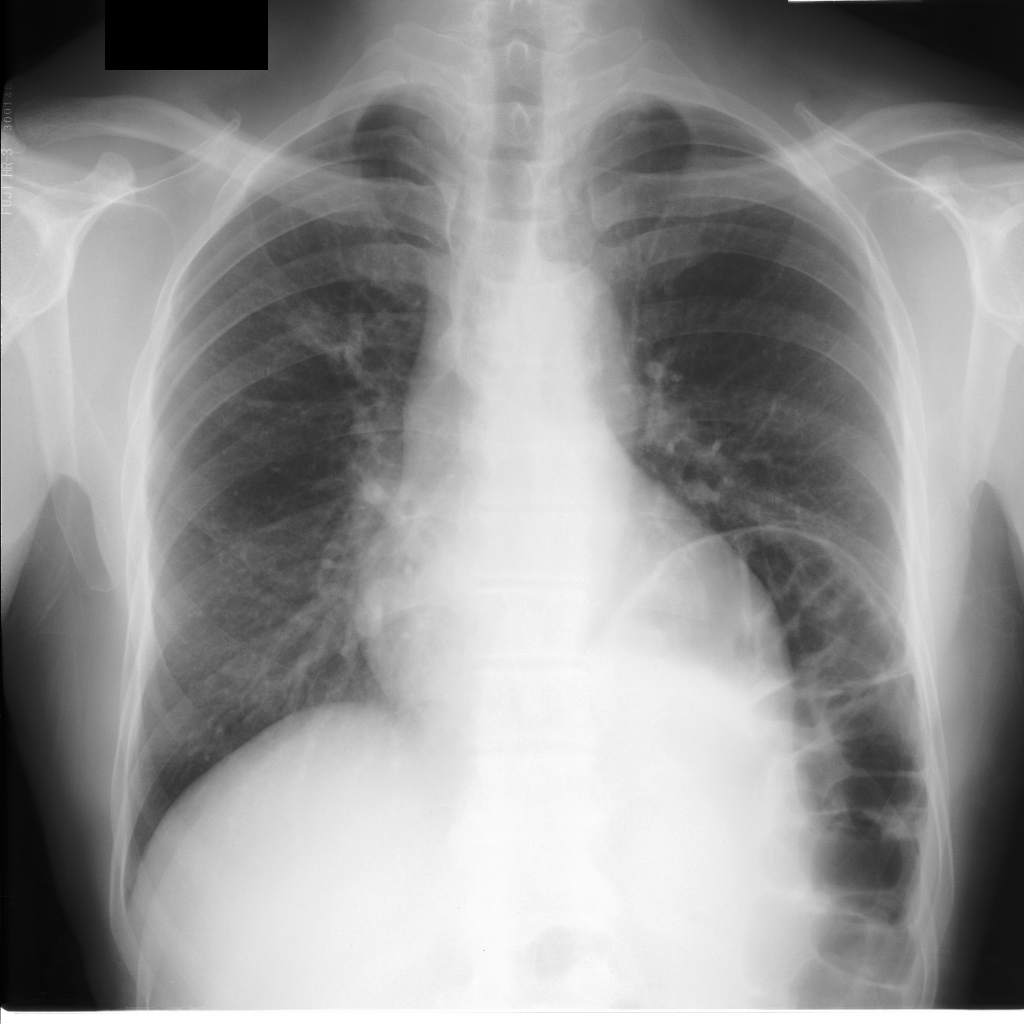}
&
\includegraphics[width=0.125\linewidth]{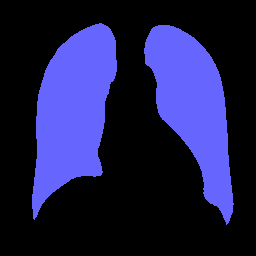}

&
\includegraphics[width=0.125\linewidth]{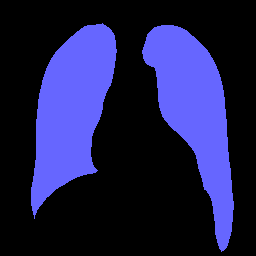}
&
\includegraphics[width=0.125\linewidth]{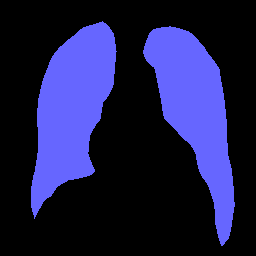}
&
\includegraphics[width=0.125\linewidth]{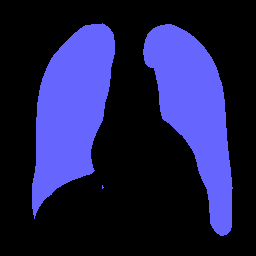}
&
\includegraphics[width=0.125\linewidth]{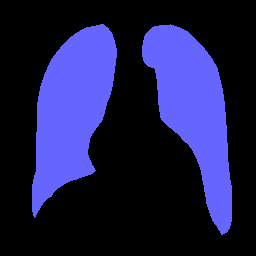}
&
\includegraphics[width=0.125\linewidth]{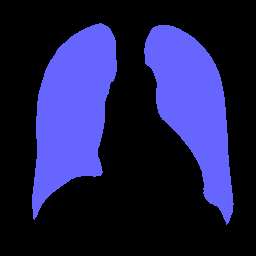}
\\
\multirow{2}{*}{\rotatebox[origin=c]{90}{NIH}}
&
\includegraphics[width=0.125\linewidth]{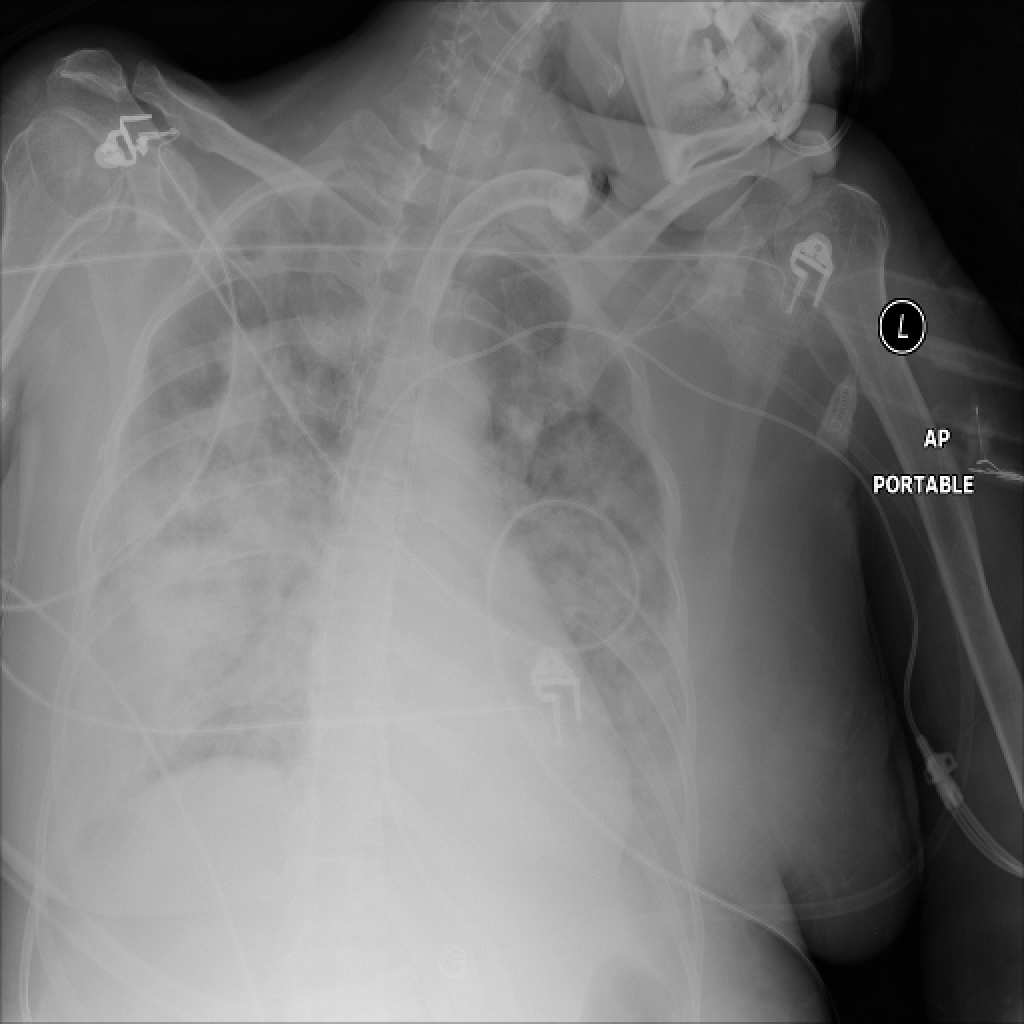}
&
\includegraphics[width=0.125\linewidth]{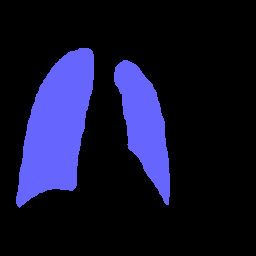}

&
\includegraphics[width=0.125\linewidth]{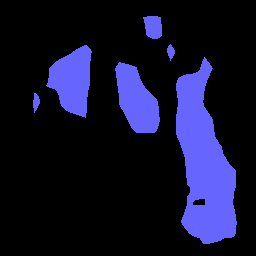}
&
\includegraphics[width=0.125\linewidth]{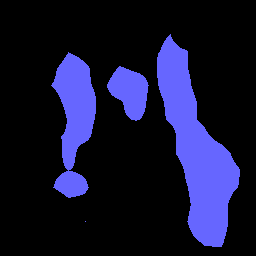}
&
\includegraphics[width=0.125\linewidth]{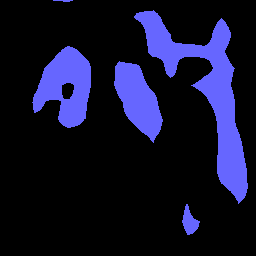}
&
\includegraphics[width=0.125\linewidth]{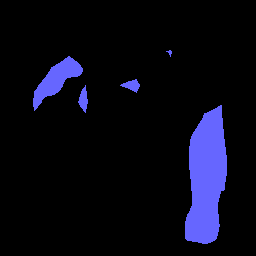}
&
\includegraphics[width=0.125\linewidth]{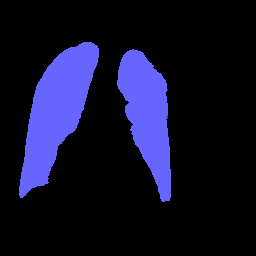}
\\
&
\includegraphics[width=0.125\linewidth]{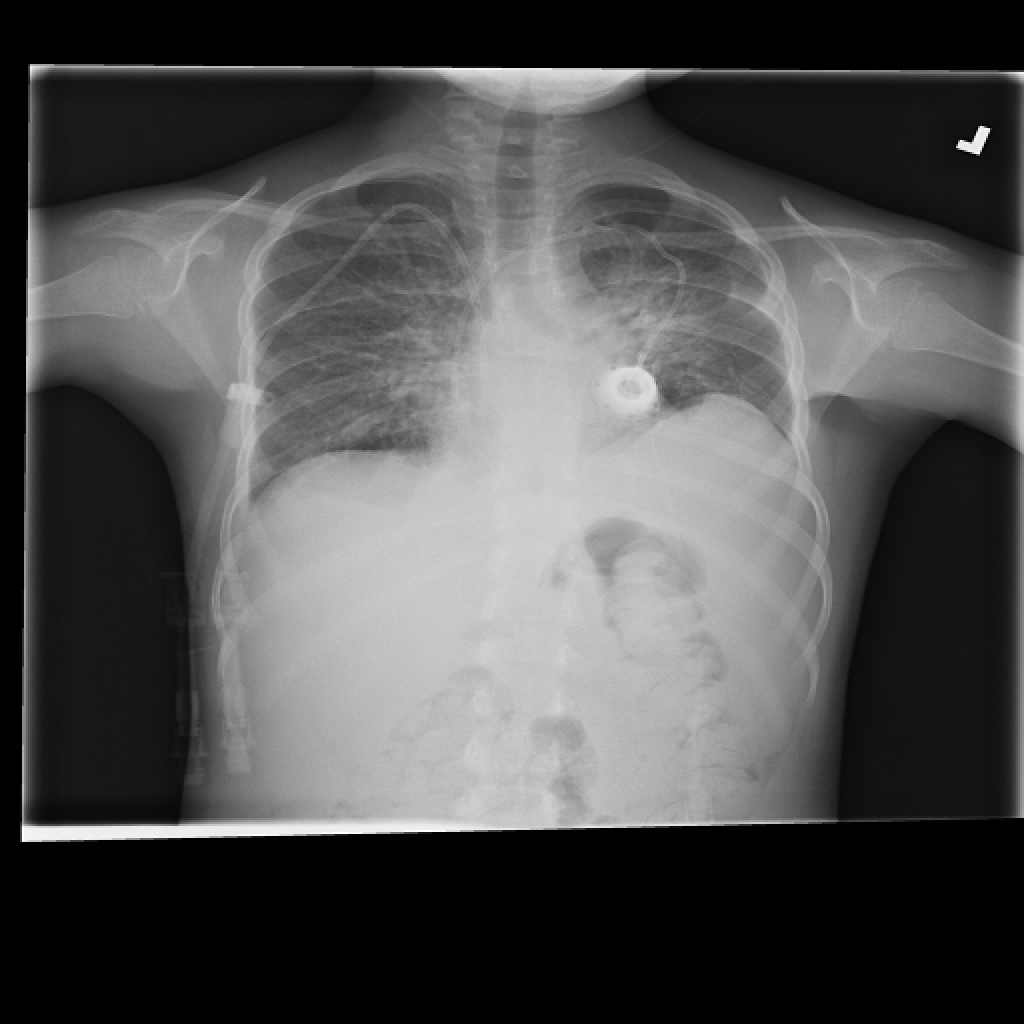}
&
\includegraphics[width=0.125\linewidth]{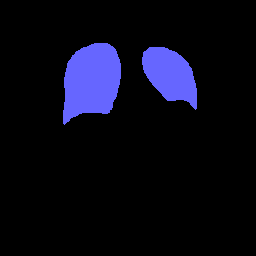}

&
\includegraphics[width=0.125\linewidth]{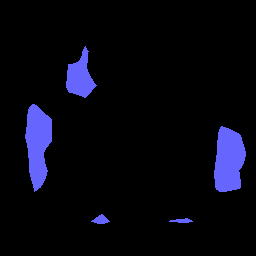}
&
\includegraphics[width=0.125\linewidth]{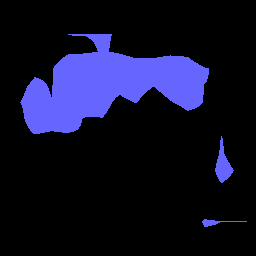}
&
\includegraphics[width=0.125\linewidth]{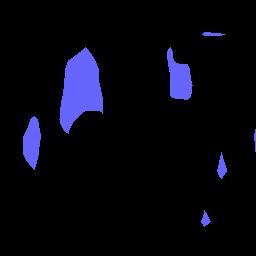}
&
\includegraphics[width=0.125\linewidth]{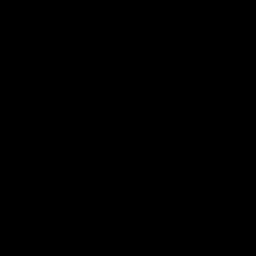}
&
\includegraphics[width=0.125\linewidth]{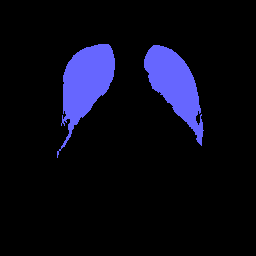}
\\
\rotatebox{90}{\scriptsize \hspace{2.5mm}NLM}
&
\includegraphics[width=0.125\linewidth]{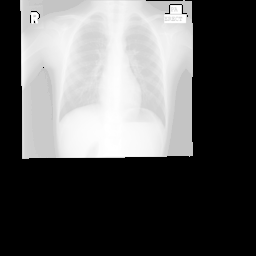}
&
\includegraphics[width=0.125\linewidth]{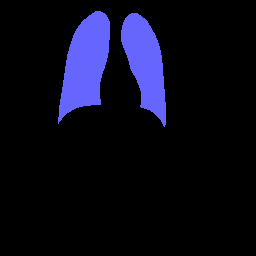}
& 
\includegraphics[width=0.125\linewidth]{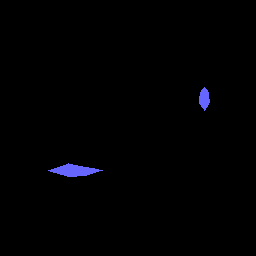}
& 
\includegraphics[width=0.125\linewidth]{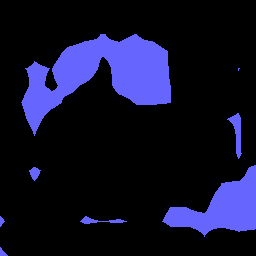}
& 
\includegraphics[width=0.125\linewidth]{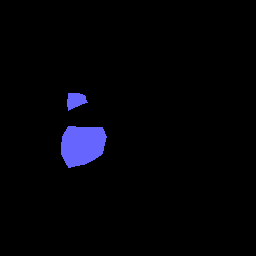}
&
\includegraphics[width=0.125\linewidth]{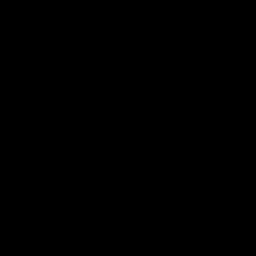}

&
\includegraphics[width=0.125\linewidth]{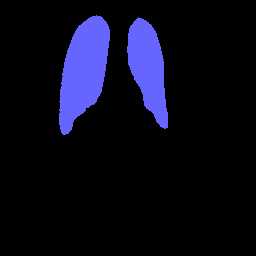}

\\
\rotatebox{90}{\scriptsize \hspace{2.5mm}SZ}
&
\includegraphics[width=0.125\linewidth]{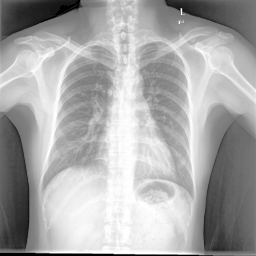}
&
\includegraphics[width=0.125\linewidth]{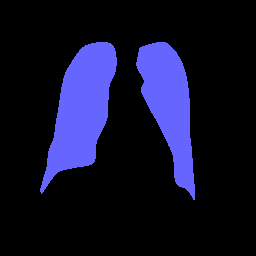}
&
\includegraphics[width=0.125\linewidth]{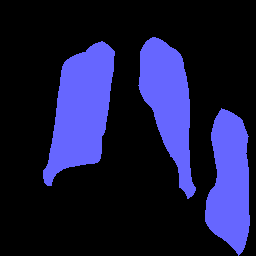}
&
\includegraphics[width=0.125\linewidth]{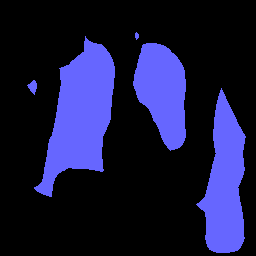}
&
\includegraphics[width=0.125\linewidth]{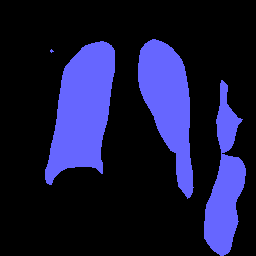}
&
\includegraphics[width=0.125\linewidth]{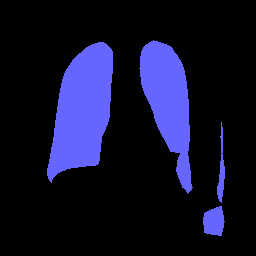}
&
\includegraphics[width=0.125\linewidth]{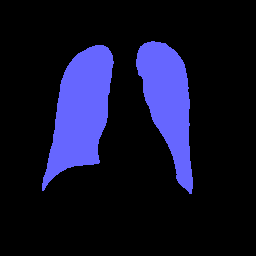}

\end{tabular}
\end{adjustbox}
\vspace{-4mm}
\caption{\footnotesize \textbf{Chest X-ray Segmentation.} Qualitative examples for both in-domain and out-of-domain datasets.}
\label{fig:chest xray segmentation}
\end{minipage}
\hspace{4.2mm}
\begin{minipage}{0.48\linewidth}
\begin{adjustbox}{center}
\footnotesize
\addtolength{\tabcolsep}{-4pt}
\begin{tabular}{cccccccc}
& image & GT & DeepLab & MT & AdvSSL & GCT & Ours \\

\rotatebox{90}{\scriptsize \hspace{-1mm}In-Domain}
&
\includegraphics[width=0.125\linewidth]{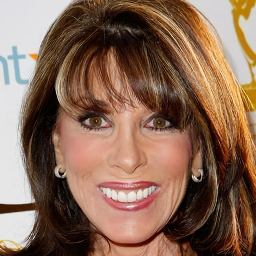}
&
\includegraphics[width=0.125\linewidth]{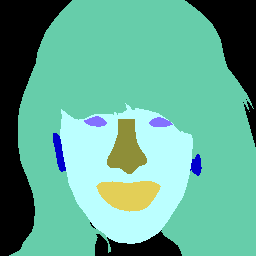}
&
\includegraphics[width=0.125\linewidth]{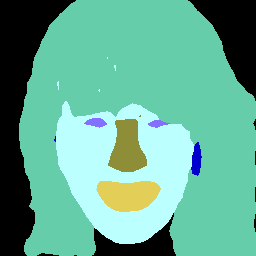}
&
\includegraphics[width=0.125\linewidth]{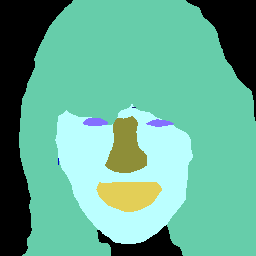}
&
\includegraphics[width=0.125\linewidth]{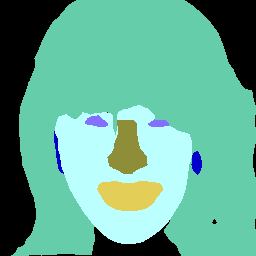}
&
\includegraphics[width=0.125\linewidth]{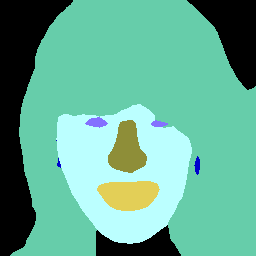}
&
\includegraphics[width=0.125\linewidth]{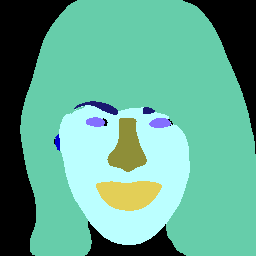}
\\
\multirow{4}{*}{\rotatebox[origin=c]{90}{Out-Domain}}
&
\includegraphics[width=0.125\linewidth]{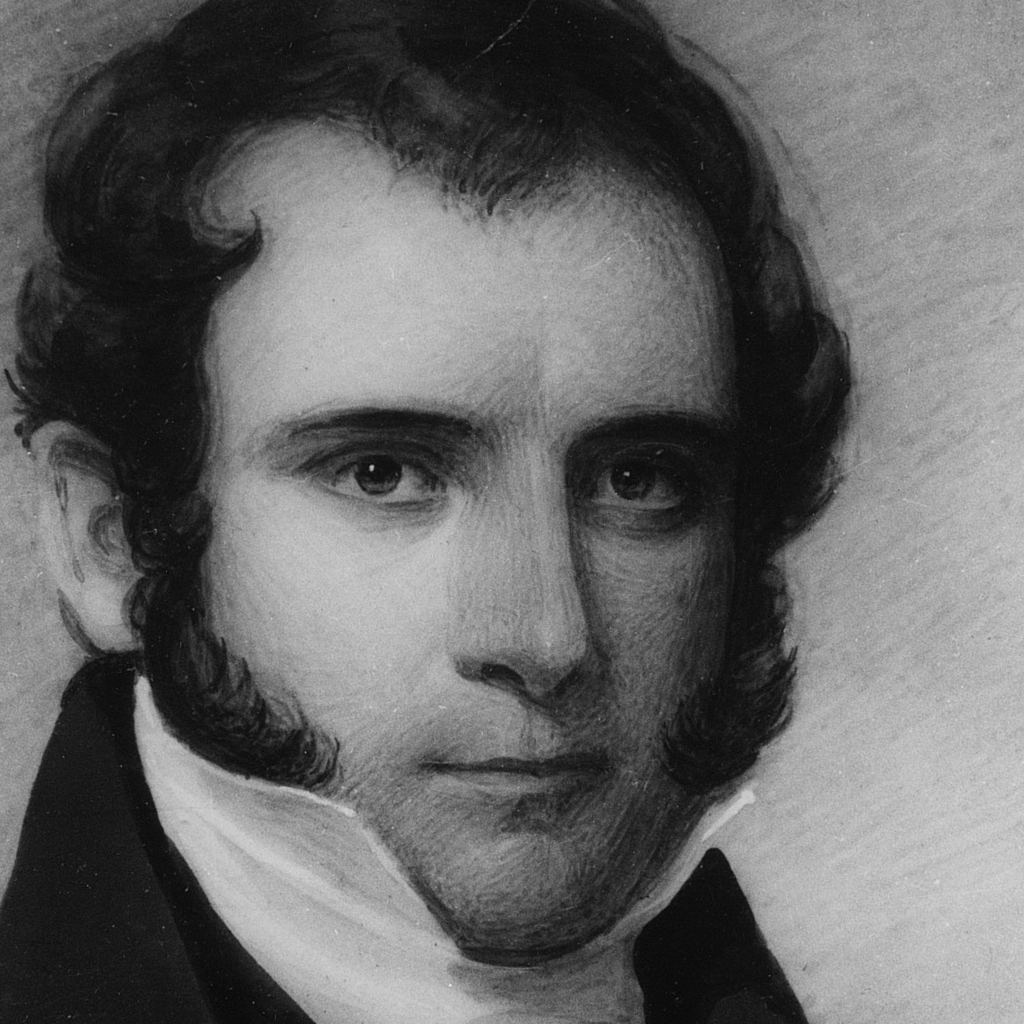}
&
\includegraphics[width=0.125\linewidth]{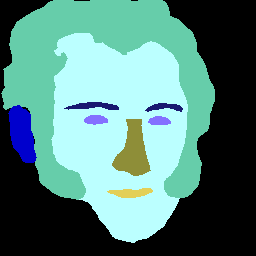}
&
\includegraphics[width=0.125\linewidth]{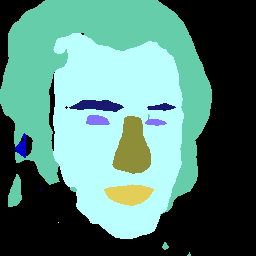}
&
\includegraphics[width=0.125\linewidth]{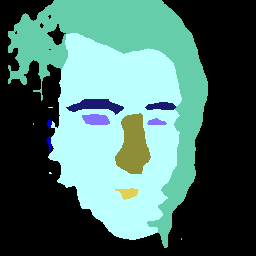}
&
\includegraphics[width=0.125\linewidth]{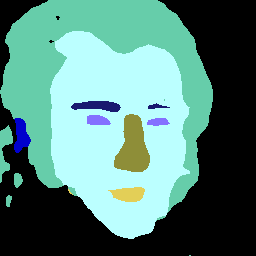}
&
\includegraphics[width=0.125\linewidth]{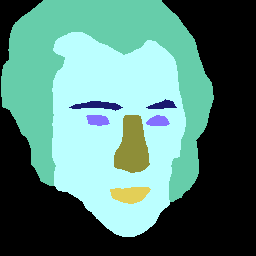}
&
\includegraphics[width=0.125\linewidth]{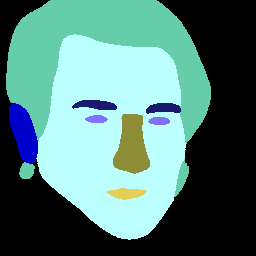}
\\
&
\includegraphics[width=0.125\linewidth]{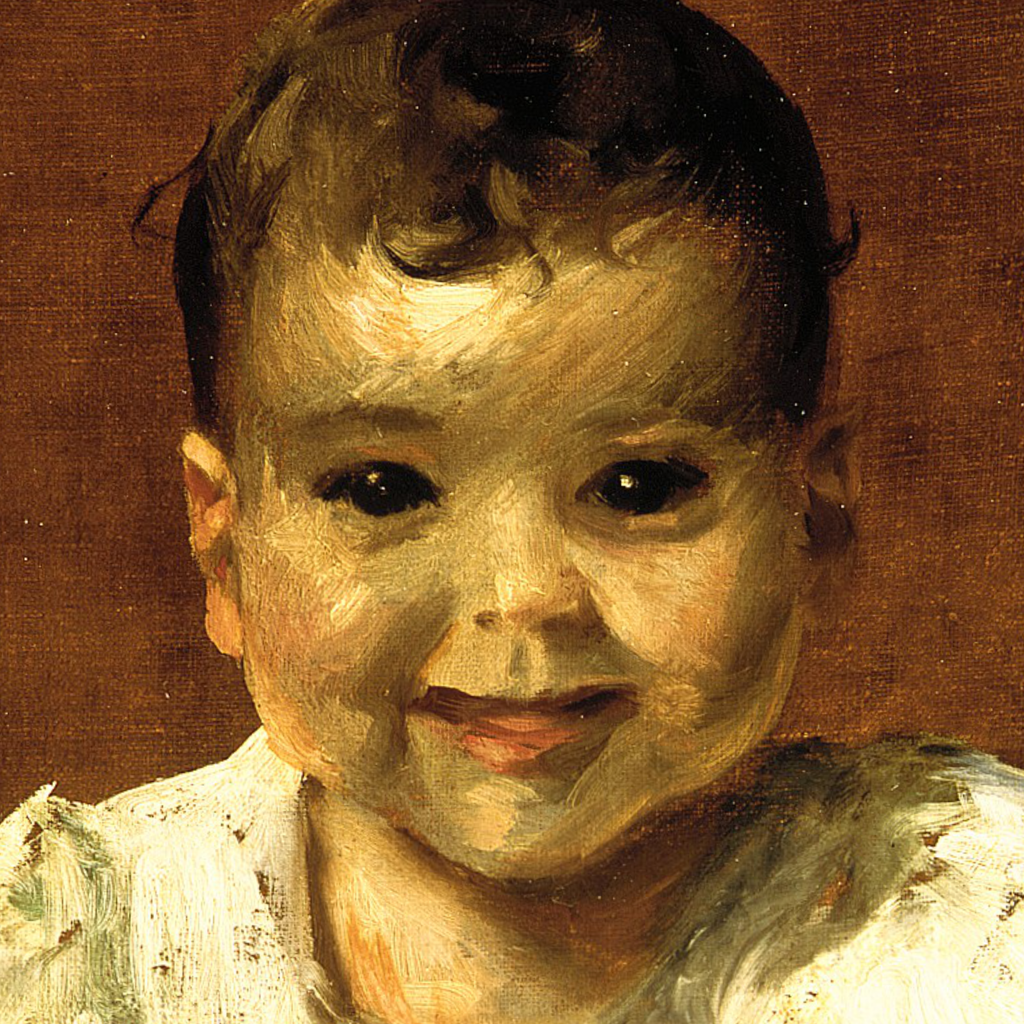}
&
\includegraphics[width=0.125\linewidth]{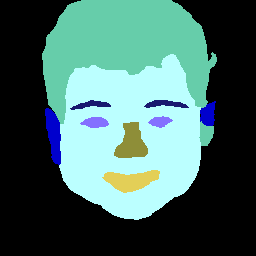}
&
\includegraphics[width=0.125\linewidth]{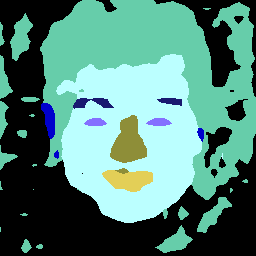}
&
\includegraphics[width=0.125\linewidth]{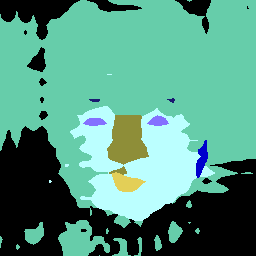}
&
\includegraphics[width=0.125\linewidth]{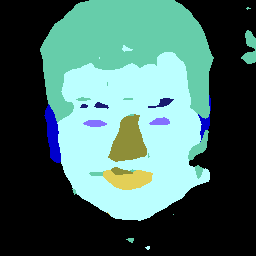}
&
\includegraphics[width=0.125\linewidth]{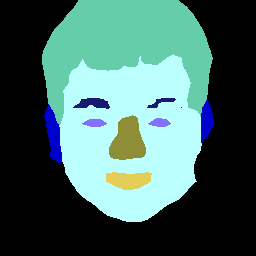}
&
\includegraphics[width=0.125\linewidth]{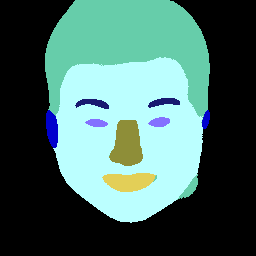}
\\
&
\includegraphics[width=0.125\linewidth]{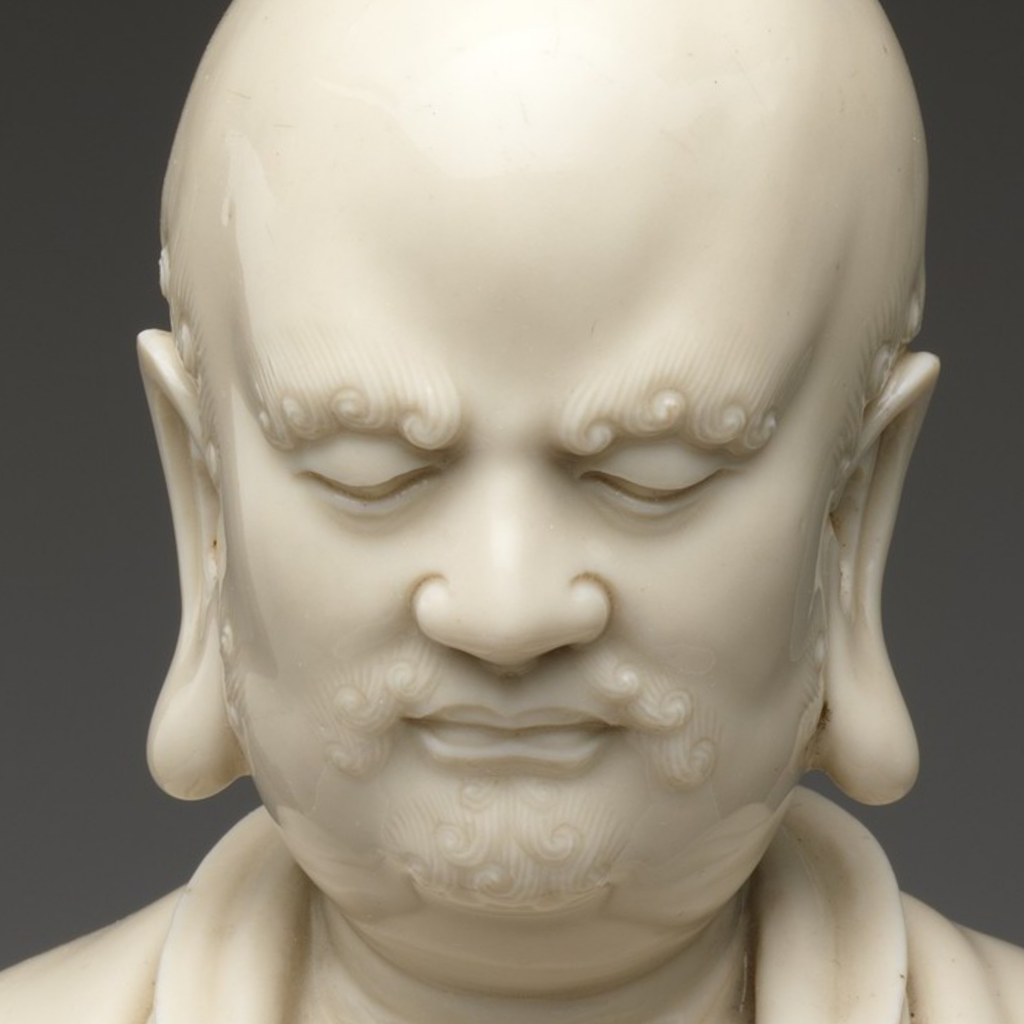}
&
\includegraphics[width=0.125\linewidth]{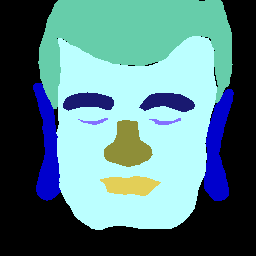}
&
\includegraphics[width=0.125\linewidth]{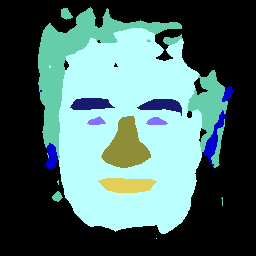}
&
\includegraphics[width=0.125\linewidth]{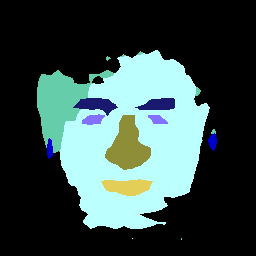}
&
\includegraphics[width=0.125\linewidth]{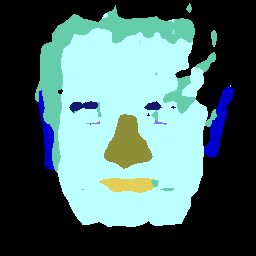}
&
\includegraphics[width=0.125\linewidth]{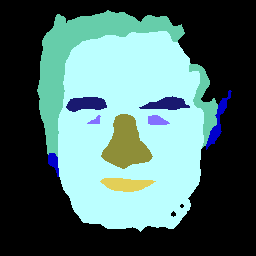}
&
\includegraphics[width=0.125\linewidth]{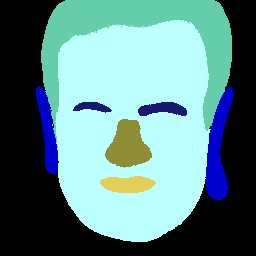}
\\
&
\includegraphics[width=0.125\linewidth]{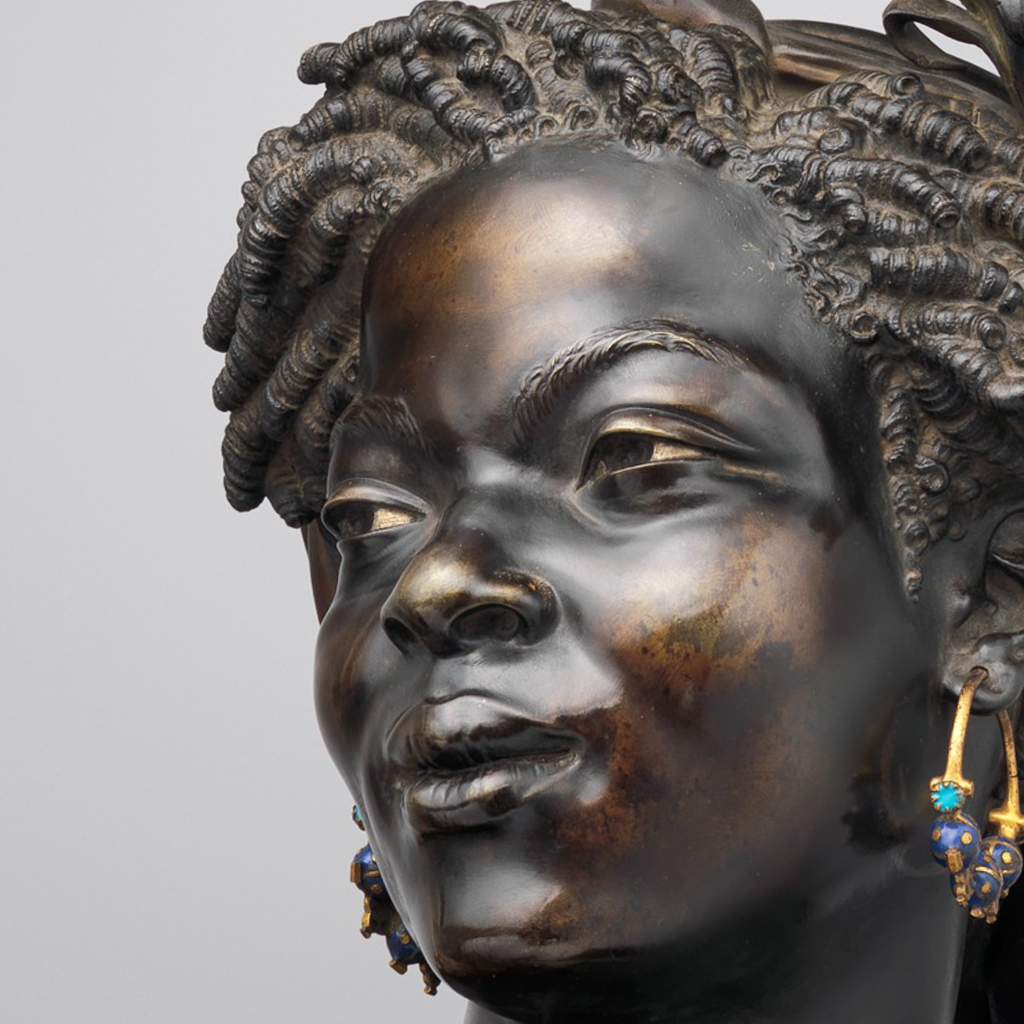}
&
\includegraphics[width=0.125\linewidth]{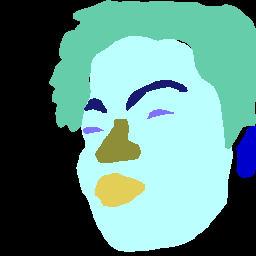}
&
\includegraphics[width=0.125\linewidth]{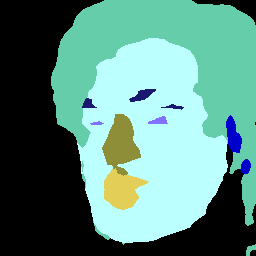}
&
\includegraphics[width=0.125\linewidth]{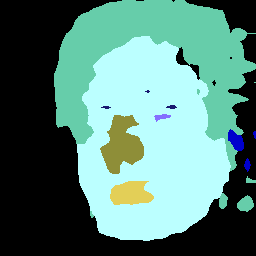}
&
\includegraphics[width=0.125\linewidth]{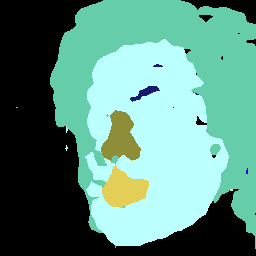}
&
\includegraphics[width=0.125\linewidth]{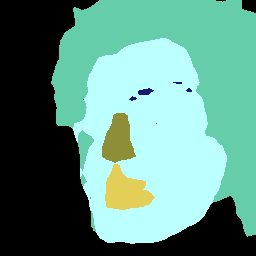}
&
\includegraphics[width=0.125\linewidth]{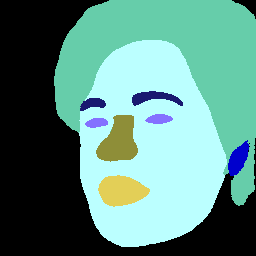}
\\
\end{tabular}
\end{adjustbox}
\vspace{-4mm}
\caption{\footnotesize \textbf{Face Parts Segmentation.} Qualitative examples for both in-domain and out-of-domain datasets.}
\label{fig:face segmentation}
\end{minipage}
\vspace{-4mm}
\end{figure*}

\textbf{Baselines.}
As baselines, we use both fully-supervised approaches, which use only the annotated subset of the data, as well as semi-supervised semantic segmentation methods, which also utilize the additional unlabeled data. With regards to fully supervised methods, the most widely used segmentation network in the medical field is \textit{U-Net}~\cite{ronneberger2015u}. Furthermore, following \cite{hung2018adversarial,mittal2019semi,ke2020guided} we compare to DeepLabV2~\cite{chen2018deeplabv2} (denoted as \textit{DeepLab} below), a stable and commonly-used architecture in the computer vision community for segmentation tasks. We also benchmark our approach against several state-of-the-art SLL methods for segmentation that have code available: the mean teacher model with transformation-consistency (\textit{MT})~\cite{tarvainen2017mean,li2020transformation}, the adversarial training-based method \cite{hung2018adversarial} (\textit{AdvSSL}), and also the recently proposed Guided Collaborative Training (\textit{GCT}) \cite{ke2020guided}.
All baselines share the same ResNet-50~\cite{he2016deep} backbone network architecture. For SSL baselines, we use the default settings as reported in the original paper. The implementations are based on the PixelSSL repository \footnote{https://github.com/ZHKKKe/PixelSSL}. 

We consider two versions of our own model. In one, we infer an image's embedding using the encoder only (denoted as \textit{Ours-NO}). In the other, we further perform optimization as described in Sec. \ref{sec:inference} (denoted as \textit{Ours}).

Further details about the datasets, evaluation metrics, and baselines can be found in the supplemental material.


\begin{table}[t!]
\vspace{-2mm}
\begin{center}
\resizebox{\linewidth}{!}{
\rowcolors{2}{white}{gray!15}
\addtolength{\tabcolsep}{-1pt}
\begin{tabular}{lccccccccc}
\toprule
\rowcolor{white}
& 
\multicolumn{2}{c}{\# Train labels: \textbf{30}} & &
\multicolumn{2}{c}{\# Train labels: \textbf{150}} &&
\multicolumn{2}{c}{\# Train labels: \textbf{1500}} \\
\cmidrule{2-3} \cmidrule{5-6} \cmidrule{8-9}
\rowcolor{white}
Method &In & MetFaces &&In & MetFaces &&In & MetFaces \\
\midrule

U-Net                                 & 0.5764 & 0.2803 && 0.6880 & 0.2803 && 0.7231 & 0.4086 \\
DeepLab                              & 0.5554 & 0.4262 && 0.6591 & 0.4988 && 0.7444 & 0.5661 \\

\midrule
MT                                   & 0.1082 & 0.1415 && 0.5857 & 0.4305 && 0.7094 & 0.5132 \\
AdvSSL                               & 0.5142 & 0.4026 && 0.6846 & 0.5029 && 0.7787 & 0.5995 \\
GCT                                  & 0.3694 & 0.3038 && 0.6403 & 0.4749 && 0.7660 & 0.5977 \\

\midrule
Ours-NO                              & 0.6473 & 0.5506 && 0.7016 & 0.5643 && 0.7123 & 0.5749 \\
Ours                                 & \textbf{0.6902} & \textbf{0.5883} && \textbf{0.7600} & \textbf{0.6336} && \textbf{0.7810} & \textbf{0.6633} \\
\bottomrule

\end{tabular}
}
\end{center}
\vspace{-6mm}
\caption{\footnotesize \textbf{Face Part Segmentation.} Numbers are mIoU. We train on CelebA and evaluate on CelebA as well as the MetFaces dataset. ``\# Train labels'' denotes the number of annotated examples used during training. Our model as well as the semi-supervised baselines additionally use 28k unlabeled CelebA data samples.} 
\label{tab:celeba-in-domain}
\vspace{-3mm}
\end{table}

\subsection{Semi-Supervised Segmentation Results}

\textbf{Chest X-ray Segmentation.} Table \ref{tab:xray-out-domain} shows our results for chest x-ray segmentation. We see that when evaluating on in-domain data, our model is on-par or better than all baselines.
When evaluating on other, out-of-domain chest x-rays, our model outperforms all baselines, both the fully supervised and semi-supervised ones, often by a large margin.
Examples of different segmentations are in Fig.~\ref{fig:chest xray segmentation}.

\begin{figure}[t!]
\vspace{-1mm}
\begin{minipage}{1\linewidth}
\centering
\begin{adjustbox}{width=0.98\linewidth}
\scriptsize
\addtolength{\tabcolsep}{-4.6pt}
\begin{tabular}{cccccccc}
& image & GT & DeepLab & MT & AdvSSL & GCT& Ours \\
\rotatebox{90}{\scriptsize \hspace{-1mm}In-Domain}
&
\includegraphics[width=0.125\linewidth]{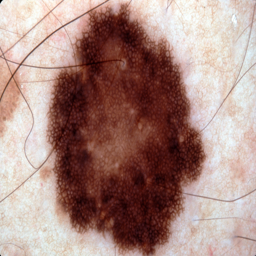}
&
\includegraphics[width=0.125\linewidth]{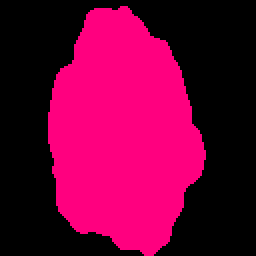}
&
\includegraphics[width=0.125\linewidth]{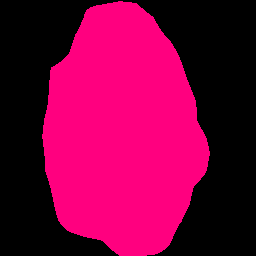}
&
\includegraphics[width=0.125\linewidth]{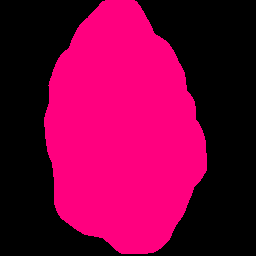}
&
\includegraphics[width=0.125\linewidth]{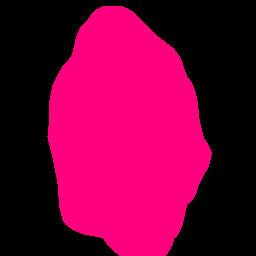}
&
\includegraphics[width=0.125\linewidth]{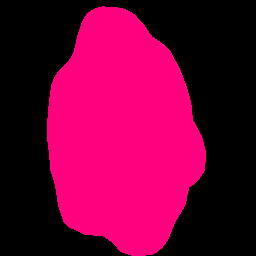}
&
\includegraphics[width=0.125\linewidth]{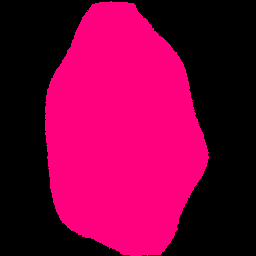}

\\
\rotatebox{90}{\scriptsize \hspace{2.7mm}PH2}
&
\includegraphics[width=0.125\linewidth]{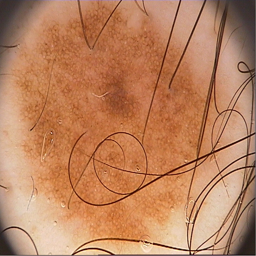}
&
\includegraphics[width=0.125\linewidth]{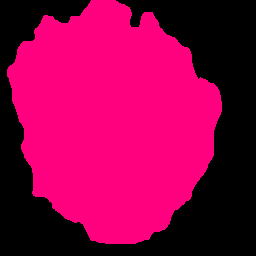}
&
\includegraphics[width=0.125\linewidth]{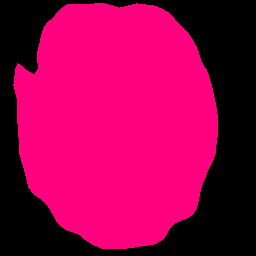}
&
\includegraphics[width=0.125\linewidth]{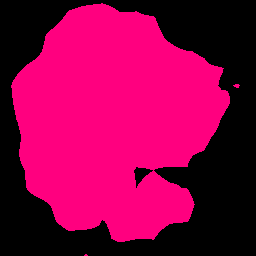}
&
\includegraphics[width=0.125\linewidth]{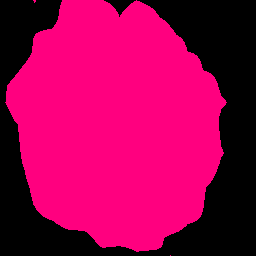}
&
\includegraphics[width=0.125\linewidth]{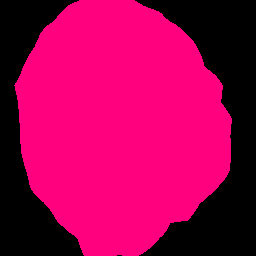}
&
\includegraphics[width=0.125\linewidth]{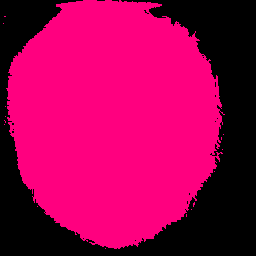}
\\
\rotatebox{90}{\scriptsize \hspace{1.2mm}dermIS}
&
\includegraphics[width=0.125\linewidth]{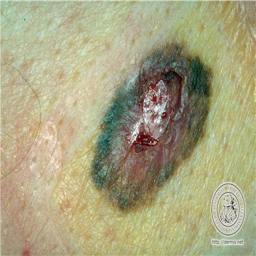}
&
\includegraphics[width=0.125\linewidth]{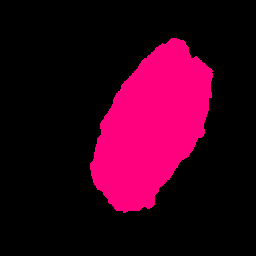}
&
\includegraphics[width=0.125\linewidth]{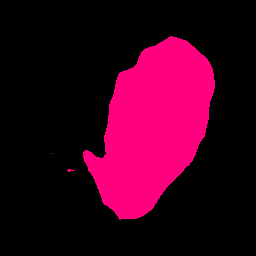}
&
\includegraphics[width=0.125\linewidth]{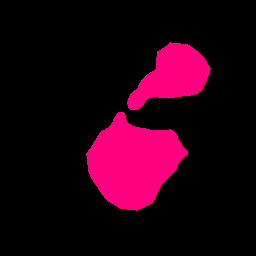}
&
\includegraphics[width=0.125\linewidth]{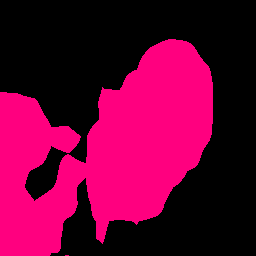}
&
\includegraphics[width=0.125\linewidth]{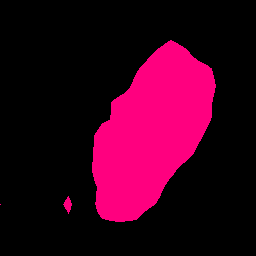}
&
\includegraphics[width=0.125\linewidth]{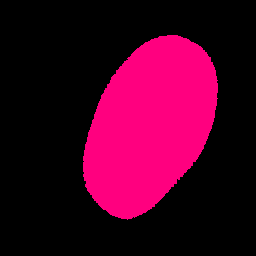}
\\
\rotatebox{90}{\scriptsize \hspace{1.2mm}dermQ.}
&
\includegraphics[width=0.125\linewidth]{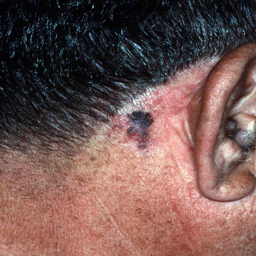}
&
\includegraphics[width=0.125\linewidth]{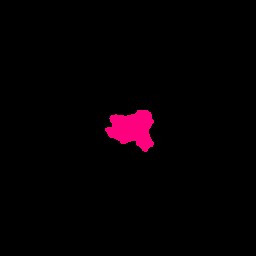}
&
\includegraphics[width=0.125\linewidth]{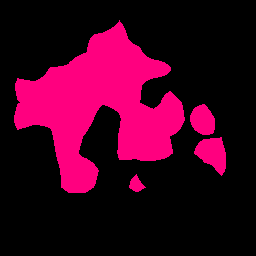}
&
\includegraphics[width=0.125\linewidth]{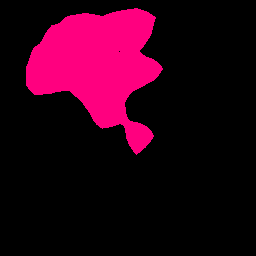}
&
\includegraphics[width=0.125\linewidth]{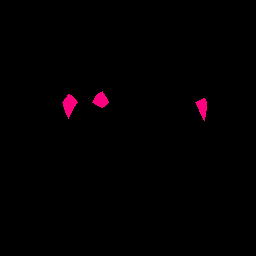}
&
\includegraphics[width=0.125\linewidth]{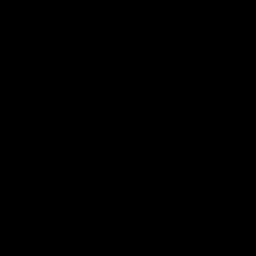}
&
\includegraphics[width=0.125\linewidth]{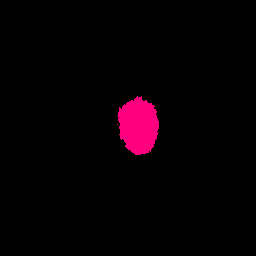}
\\

\end{tabular}
\end{adjustbox}
\vspace{-4.5mm}
\caption{\footnotesize \textbf{Skin Lesion Segmentation.} Qualitative examples for both in-domain and out-of-domain datasets.}
\label{fig:qual_skin}
\end{minipage}
\vspace{-5.5mm}
\end{figure}

\textbf{Skin Lesion Segmentation.} Table \ref{tab:isic-out-domain} presents the results for skin lesion segmentation (also see Figure \ref{fig:qual_skin} for visualizations). The gap between our method and the baselines is even more pronounced. We consistently outperform all baseslines, both supervised and semi-supervised ones as well as both in-domain and during evaluation on out-of-domain data. 

\newcommand\hh{1.12cm}
\begin{figure*}[t!]
\vspace{-2mm}
\begin{minipage}{0.55\linewidth}
\addtolength{\tabcolsep}{-4.5pt}
\begin{tabular}{cccccccc}
\includegraphics[height=\hh]{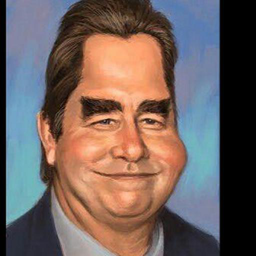}
&
\includegraphics[height=\hh]{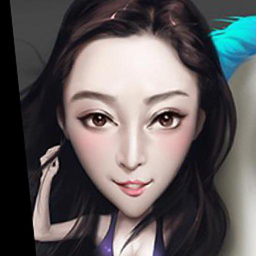}
&
\includegraphics[height=\hh]{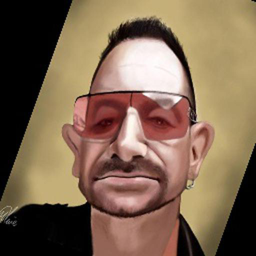}
&
\includegraphics[height=\hh]{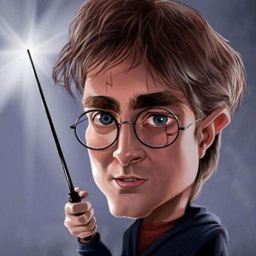}
&
\includegraphics[height=\hh]{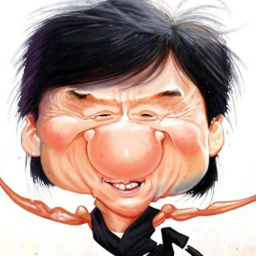}
&
\includegraphics[height=\hh]{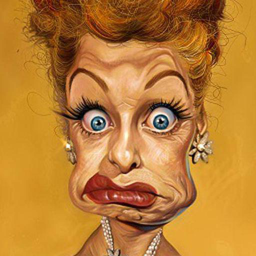}
&
\includegraphics[height=\hh]{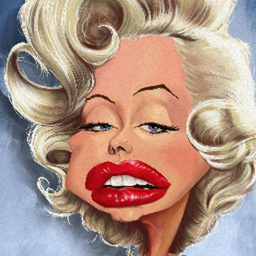}
&
\includegraphics[height=\hh]{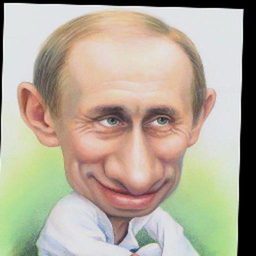}
\\

\includegraphics[height=\hh]{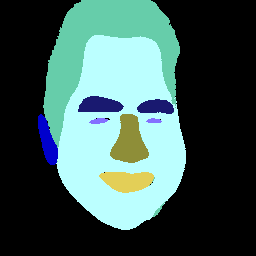}
&
\includegraphics[height=\hh]{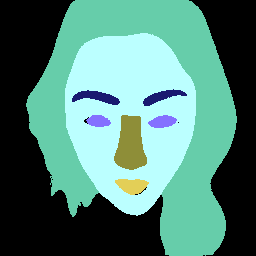}
&
\includegraphics[height=\hh]{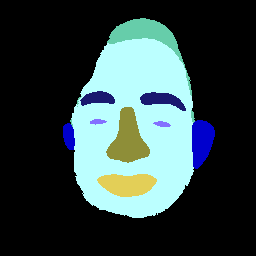}
&
\includegraphics[height=\hh]{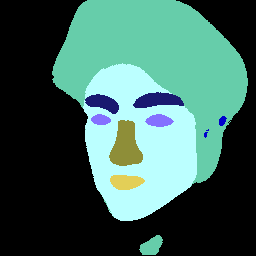}
&
\includegraphics[height=\hh]{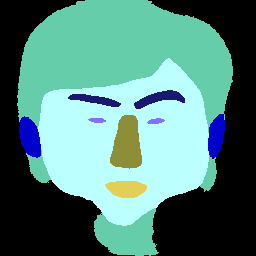}
&
\includegraphics[height=\hh]{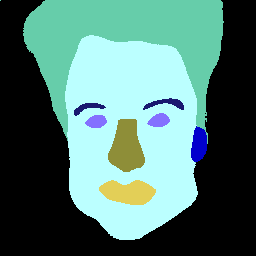}
&
\includegraphics[height=\hh]{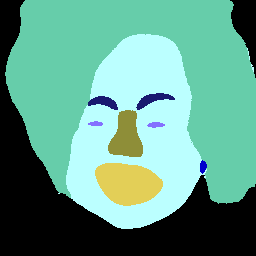}
&
\includegraphics[height=\hh]{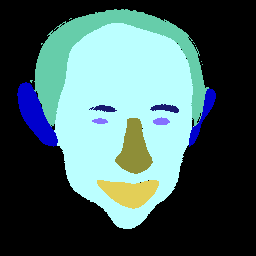}\\
\end{tabular}
\end{minipage}
\hspace{2mm}
\begin{minipage}{0.46\linewidth}
\addtolength{\tabcolsep}{-4.5pt}
\begin{tabular}{cccccc}
\includegraphics[height=\hh,trim=0 0 0 0,clip]{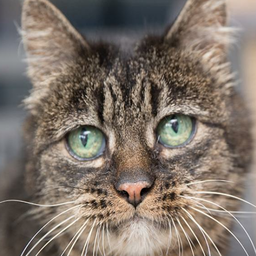}
&
\includegraphics[height=\hh]{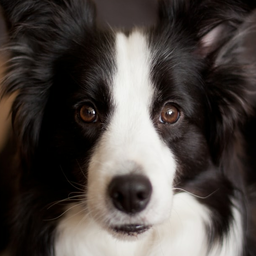}
&
\includegraphics[height=\hh]{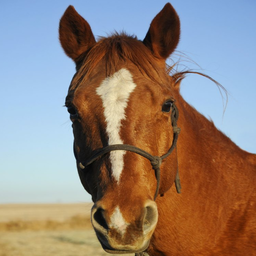}
&
\includegraphics[height=\hh]{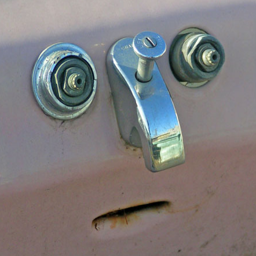}
&
\includegraphics[height=\hh]{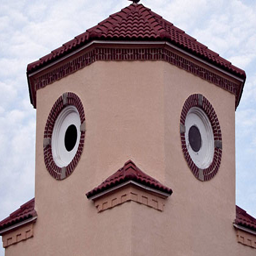}
&
\includegraphics[height=\hh]{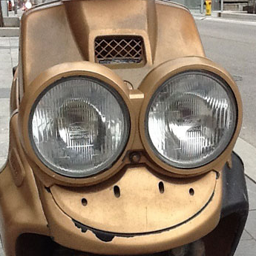}
\\
\includegraphics[height=\hh]{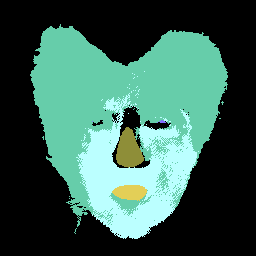}
&
\includegraphics[height=\hh]{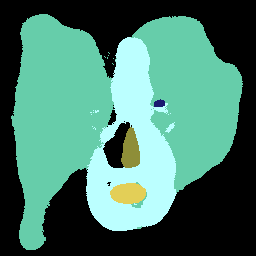}
&
\includegraphics[height=\hh]{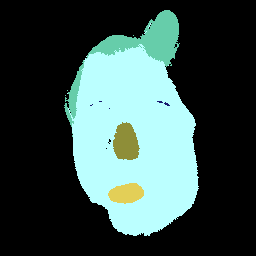}
&
\includegraphics[height=\hh]{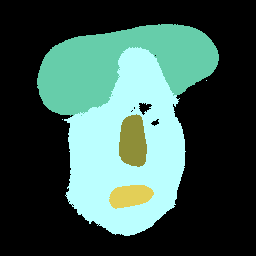}
&
\includegraphics[height=\hh]{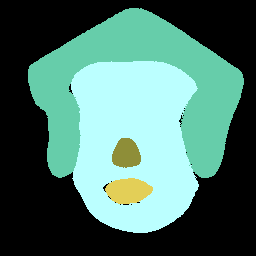}
&
\includegraphics[height=\hh]{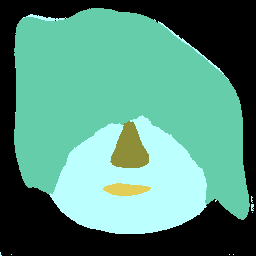}
\\
\end{tabular}
\end{minipage}
\vspace{-4mm}
\caption{\footnotesize \textbf{Extreme Out-Of-Domain Segmentation.} Results on images with a large visual gap to CelebA, on which our model was trained.}
\label{fig:face extreme}
\end{figure*}

\textbf{Face Part Segmentation.} We observe similar results for face part segmentation, where we outperform all baselines (see Table~\ref{tab:celeba-in-domain} and Figure \ref{fig:face segmentation}). In particular for out-of-domain segmentation on the MetFaces data set, we find that we beat the other methods by a large margin. Since our method is designed with semi-supervised training in mind, we trained the models with a limited number of annotations. For reference, we additionally trained a DeepLab model with all 28k mask annotations of the CelebA dataset. This model achieves $0.7945$ mIoU when evaluated on CelebA test data and $0.6415$ mIoU when evaluated on MetFaces test data. Comparing to Table \ref{tab:celeba-in-domain}, this means that our method, using only 1.5k labels, even outperforms a modern DeepLab model that was trained with all available 28k labels when evaluated on out-of-domain MetFaces data. This is a testament to our model's strong generalization and efficient semi-supervised training capabilities.

Encouraged by these results we experiment with evaluating our CelebA model also on more extreme out-of-domain images. We test our model on cartoons, faces of animals and even non-face images that exhibit face-like features (see Figure \ref{fig:face extreme}). Qualitatively, we observe that we can generate reasonable segmentations even for these extreme out-of-domain examples, a feat that hasn't been demonstrated before, to the best of our knowledge.

\textbf{CT-MRI Transfer.} Having observed that our model demonstrates very strong generalization properties in the visual domain, we explore an additional far-out-of-domain problem in medical image analysis: We train our segmentation method on CT images and evaluate on MR images for liver segmentation. Our results in Table~\ref{tab:liver-out-domain} demonstrate that our model outperforms the chosen supervised baselines on this very challenging out-of-domain segmentation task by a large margin. Details about this additional experiment are in the supplemental material.

We attribute our model's strong generalization performance in the semi-supervised setting to its design as a fully generative model. Our experimental results validate our assumptions and motivations discussed in Sec. \ref{sec:methodmotivation}. We also find that we generally obtain better results when refining an image's inferred embedding via optimization, as described in Sec. \ref{sec:inference}, instead of directly using the encoder prediction.



\subsection{Value of Data \& Training with Generated Data}

We conduct an ablation study on the amount of unlabeled and labeled data used in our method. Traditionally, labeled data is considered more valuable than unlabeled data but there is no clear understanding of how many unlabeled data points boost performance as much as a labeled data sample. We measure the value of data in terms of segmentation performance (mIoU). In Table \ref{tab:label-unlabel}, we report performance for different amounts of labeled and unlabeled data used during training.
Interestingly, we observe that the performance with 1500 labeled and 3K unlabeled data is almost equivalent to 150 labeled and 28K unlabeled data samples.

\definecolor{orange}{rgb}{1,0.8,0.8}
\definecolor{lblue}{rgb}{0.75,0.85,1}
\begin{table}[t!]
\vspace{-5mm}
\begin{minipage}{0.495\linewidth}
\begin{center}
\begin{adjustbox}{width=1\linewidth}
\footnotesize
\addtolength{\tabcolsep}{-4pt}
\begin{tabular}{|p{1.1mm}l|c|c|c|}
\hline
& & \multicolumn{3}{c|}{Unlabeled}\\
& &  3K & 10K & 28K \\ 
\hline
\multirow{2}{*}{\rotatebox[origin=c]{90}{Labeled}} & 30      & 0.6786 & 0.6845 & {\cellcolor{orange} 0.6902} \\
 & 150     & {\cellcolor{orange}0.7046} & 0.7438 & {\cellcolor{lblue} 0.7600} \\
& 1500    & {\cellcolor{lblue} 0.7566} & 0.7710 & 0.7810 \\
\hline
\end{tabular}
\end{adjustbox}
\end{center}
\end{minipage}
\begin{minipage}{0.49\linewidth}
\begin{center}
\begin{adjustbox}{width=\linewidth}
\scriptsize
\addtolength{\tabcolsep}{-4pt}
\begin{tabular}{|p{1.1mm}l|c|c|c|}
\hline
& & \multicolumn{3}{c|}{Unlabeled}\\   
 & &  3K & 10K & 28K \\ 
\hline
\multirow{2}{*}{\rotatebox[origin=c]{90}{Labeled}} & 30      & 0.5410 & 0.5799 & \cellcolor{orange}{0.5883} \\
& 150     & \cellcolor{orange}{0.5871} & 0.6152 & \cellcolor{lblue}{0.6336} \\
& 1500    & 0.6011 & \cellcolor{lblue}{0.6204} & 0.6633 \\
\hline
\end{tabular}
\end{adjustbox}
\end{center}
\end{minipage}
\footnotesize
\begin{tabular}{p{0.47\linewidth}p{0.49\linewidth}}
\hspace{1.5mm}(a) CelebA-Mask (In-Domain) & \hspace{0mm}(b) MetFaces-40 (Out-Domain)\\
\end{tabular}
\vspace{-4.5mm}
\caption{\footnotesize \textbf{Ablation Study on Number of Labeled vs Unlabeled Examples.} Numbers are mIoU. Entries marked with red or blue color roughly correspond to each other, \ie 30 labeled and 28k unlabeled results in similar performance as 150 labeled and 3k unlabeled examples.}
\label{tab:label-unlabel}
\vspace{-3mm}
\end{table}

Simulation is often used to directly generate annotated synthetic data, reducing the need for expensive manual labelling. However, it is unclear to which degree synthetic data is useful for downstream tasks, due to the domain gap between simulated and real data. We conduct another experiment to evaluate the value of synthetic labeled data. 
Since our method models the joint image-label distribution, we can also use our model to generate a large amount of synthetic but annotated images. These can then be used to train a regular segmentation network in a fully-supervised, discriminative manner.
Specifically, we sample 20k synthetic face images and their pixel-wise labels, using two different sampling strategies, and then train DeepLabV2 segmentation models with this data. 
Our results, presented in Table~\ref{tab:sim-data}, show that high quality synthetic data is useful for the downstream task. We explored different strategies on how to sample and use the data and find that they all 
beat the baseline that was trained with real data only. However, this approach is sensitive to the sampling strategy used to generate the data.
Importantly, we also observe that directly doing segmentation with the generative model, as proposed in this paper, performs best by a large margin. However, doing segmentation with the generative model requires test-time optimization and is thus not suitable for real-time applications. Speed-ups are  future work.

\begin{table}[t!]
\vspace{-5.5mm}
\begin{center}
\renewcommand{\arraystretch}{0.6}
\resizebox{0.7\linewidth}{!}{
\rowcolors{2}{white}{gray!15}
\footnotesize
\addtolength{\tabcolsep}{3pt}
\begin{tabular}{lcc}
\toprule
&\multicolumn{2}{c}{Dataset} \\
\cmidrule{2-3}
\rowcolor{white}
Method & CelebA & MetFaces \\

\midrule
DeepLab-real      & 0.6591 & 0.4988 \\
\midrule

Ours-\textit{sim}-\textit{tru}      & 0.6829 & 0.5137 \\
Ours-\textit{mix}-\textit{tru}      & 0.7159 & 0.5498 \\
Ours-\textit{sim}-\textit{div}      & 0.7051 & 0.5569 \\
Ours-\textit{mix}-\textit{div}      & 0.7192 & 0.5656 \\
\midrule
Ours              & 0.7600 & 0.6336 \\

\bottomrule
\end{tabular}
}
\end{center}
\renewcommand{\arraystretch}{1.0}
\vspace{-6mm}
\caption{\footnotesize \textbf{Synthesize Annotated Images to Train a Task Model vs Our Method.} Numbers are mIoU. \textit{DeepLab-real} denotes supervised training of a DeepLab model using 150 labeled real examples. \textit{Ours-sim} denotes training DeepLab using only the 20k synthetic dataset. \textit{Ours-mix} means training DeepLab using both the synthetic and 150 labeled real examples. \textit{div} denotes sampling without applying the truncation trick~\cite{karras2019style}, which results in more diverse but less visually appealing images; \textit{tru} means applying the truncation trick with factor of $0.7$. \textit{Ours} denotes performing segmentation directly with our generative segmentation method.}
\label{tab:sim-data}
\vspace{-4mm}
\end{table}

\vspace{-2mm}
\section{Conclusion}
\vspace{-1mm}
In this paper, we propose a fully generative approach to semantic segmentation, based on StyleGAN2, that naturally allows for semi-supervised training and shows very strong generalization capabilities. We validate our method in the medical domain, where annotation can be particularly expensive and where models need to transfer, for example, between different imaging sensors. 
Quantitatively, we significantly outperform available strong baselines in- as well as out-of-domain. To showcase our method's versatility, we perform additional experiments on face part segmentation. We find that our model generalizes to paintings, sculptures and cartoons. Interestingly, it produces plausible segmentations even on extreme-out-of-domain examples, such as animal faces. We attribute the model's remarkable generalization capabilities to its design as a fully generative model.

{\small
\bibliographystyle{ieee_fullname}
\bibliography{egbib}

\begin{thebibliography}{10}\itemsep=-1pt

\bibitem{abdal2019image2stylegan}
Rameen Abdal, Yipeng Qin, and Peter Wonka.
\newblock Image2stylegan: How to embed images into the stylegan latent space?
\newblock In {\em Proceedings of the IEEE International Conference on Computer
  Vision}, pages 4432--4441, 2019.

\bibitem{bai2017semi}
Wenjia Bai, Ozan Oktay, Matthew Sinclair, Hideaki Suzuki, Martin Rajchl,
  Giacomo Tarroni, Ben Glocker, Andrew King, Paul~M Matthews, and Daniel
  Rueckert.
\newblock Semi-supervised learning for network-based cardiac mr image
  segmentation.
\newblock In {\em International Conference on Medical Image Computing and
  Computer-Assisted Intervention}, pages 253--260. Springer, 2017.

\bibitem{bau2019semantic}
David Bau, Hendrik Strobelt, William Peebles, Jonas Wulff, Bolei Zhou, Jun-Yan
  Zhu, and Antonio Torralba.
\newblock Semantic photo manipulation with a generative image prior.
\newblock {\em ACM Trans. Graph.}, 38(4), July 2019.

\bibitem{bau2019seeing}
D. {Bau}, J. {Zhu}, J. {Wulff}, W. {Peebles}, B. {Zhou}, H. {Strobelt}, and A.
  {Torralba}.
\newblock Seeing what a gan cannot generate.
\newblock In {\em 2019 IEEE/CVF International Conference on Computer Vision
  (ICCV)}, pages 4501--4510, 2019.

\bibitem{bau2018gan}
David Bau, Jun-Yan Zhu, Hendrik Strobelt, Bolei Zhou, Joshua~B Tenenbaum,
  William~T Freeman, and Antonio Torralba.
\newblock Gan dissection: Visualizing and understanding generative adversarial
  networks.
\newblock {\em arXiv preprint arXiv:1811.10597}, 2018.

\bibitem{Berthelot2020ReMixMatch}
David Berthelot, Nicholas Carlini, Ekin~D. Cubuk, Alex Kurakin, Kihyuk Sohn,
  Han Zhang, and Colin Raffel.
\newblock Remixmatch: Semi-supervised learning with distribution matching and
  augmentation anchoring.
\newblock In {\em International Conference on Learning Representations}, 2020.

\bibitem{berthelot2019mixmatch}
David Berthelot, Nicholas Carlini, Ian Goodfellow, Nicolas Papernot, Avital
  Oliver, and Colin Raffel.
\newblock Mixmatch: A holistic approach to semi-supervised learning.
\newblock In {\em NeurIPS}, 2019.

\bibitem{beyer2019s4l}
L. {Beyer}, X. {Zhai}, A. {Oliver}, and A. {Kolesnikov}.
\newblock S4l: Self-supervised semi-supervised learning.
\newblock In {\em 2019 IEEE/CVF International Conference on Computer Vision
  (ICCV)}, pages 1476--1485, 2019.

\bibitem{bilic2019liver}
Patrick Bilic, Patrick~Ferdinand Christ, Eugene Vorontsov, Grzegorz Chlebus,
  Hao Chen, Qi Dou, Chi-Wing Fu, Xiao Han, Pheng-Ann Heng, J{\"u}rgen Hesser,
  et~al.
\newblock The liver tumor segmentation benchmark (lits).
\newblock {\em arXiv preprint arXiv:1901.04056}, 2019.

\bibitem{Brock2017neural}
Andrew Brock, Theodore Lim, James~M. Ritchie, and Nick Weston.
\newblock Neural photo editing with introspective adversarial networks.
\newblock In {\em 5th International Conference on Learning Representations,
  {ICLR} 2017, Toulon, France, April 24-26, 2017, Conference Track
  Proceedings}. OpenReview.net, 2017.

\bibitem{chen2018deeplabv2}
Liang-Chieh Chen, George Papandreou, Iasonas Kokkinos, Kevin Murphy, and Alan~L
  Yuille.
\newblock Deeplab: Semantic image segmentation with deep convolutional nets,
  atrous convolution, and fully connected crfs.
\newblock {\em IEEE transactions on pattern analysis and machine intelligence},
  40(4):834—848, April 2018.

\bibitem{chen2017rethinking}
Liang-Chieh Chen, George Papandreou, Florian Schroff, and Hartwig Adam.
\newblock Rethinking atrous convolution for semantic image segmentation.
\newblock {\em arXiv preprint arXiv:1706.05587}, 2017.

\bibitem{chen2020big}
Ting Chen, Simon Kornblith, Kevin Swersky, Mohammad Norouzi, and Geoffrey
  Hinton.
\newblock Big self-supervised models are strong semi-supervised learners.
\newblock In {\em NeurIPS}, 2020.

\bibitem{codella2019skin}
Noel Codella, Veronica Rotemberg, Philipp Tschandl, M~Emre Celebi, Stephen
  Dusza, David Gutman, Brian Helba, Aadi Kalloo, Konstantinos Liopyris, Michael
  Marchetti, et~al.
\newblock Skin lesion analysis toward melanoma detection 2018: A challenge
  hosted by the international skin imaging collaboration (isic).
\newblock {\em arXiv preprint arXiv:1902.03368}, 2019.

\bibitem{collins2020editing}
Edo Collins, Raja Bala, Bob Price, and Sabine Süsstrunk.
\newblock Editing in style: Uncovering the local semantics of gans.
\newblock {\em arXiv preprint arXiv:2004.14367}, 2020.

\bibitem{creswell2019inverting}
A. {Creswell} and A.~A. {Bharath}.
\newblock Inverting the generator of a generative adversarial network.
\newblock {\em IEEE Transactions on Neural Networks and Learning Systems},
  30(7):1967--1974, 2019.

\bibitem{metasim20}
Jeevan Devaranjan, Amlan Kar, and Sanja Fidler.
\newblock Meta-sim2: Unsupervised learning of scene structure for synthetic
  data generation.
\newblock In {\em ECCV}, 2020.

\bibitem{doersch2015}
Carl Doersch, Abhinav Gupta, and Alexei~A. Efros.
\newblock Unsupervised visual representation learning by context prediction.
\newblock In {\em Proceedings of the 2015 IEEE International Conference on
  Computer Vision (ICCV)}, ICCV '15, page 1422–1430, USA, 2015. IEEE Computer
  Society.

\bibitem{donahue2016adversarial}
Jeff Donahue, Philipp Kr{\"a}henb{\"u}hl, and Trevor Darrell.
\newblock Adversarial feature learning.
\newblock {\em arXiv preprint arXiv:1605.09782}, 2016.

\bibitem{dumoulin2017adversarially}
Vincent Dumoulin, Ishmael Belghazi, Ben Poole, Alex Lamb, Mart{\'{\i}}n
  Arjovsky, Olivier Mastropietro, and Aaron~C. Courville.
\newblock Adversarially learned inference.
\newblock In {\em 5th International Conference on Learning Representations,
  {ICLR} 2017, Toulon, France, April 24-26, 2017, Conference Track
  Proceedings}. OpenReview.net, 2017.

\bibitem{fang2019co}
Tiantian Fang and Alexander Schwing.
\newblock Co-generation with gans using ais based hmc.
\newblock In {\em Advances in Neural Information Processing Systems}, pages
  5808--5819, 2019.

\bibitem{French2020}
Geoffrey French, Samuli Laine, Timo Aila, Michal Mackiewicz, and Graham
  Finlayson.
\newblock Semi-supervised semantic segmentation needs strong, varied
  perturbations.
\newblock In {\em BMVC}, 2020.

\bibitem{french2020milking}
Geoff French, Avital Oliver, and Tim Salimans.
\newblock Milking cowmask for semi-supervised image classification.
\newblock {\em arXiv preprint arXiv:2003.12022}, 2020.

\bibitem{gidaris2018unsupervised}
Spyros Gidaris, Praveer Singh, and Nikos Komodakis.
\newblock Unsupervised representation learning by predicting image rotations.
\newblock In {\em International Conference on Learning Representations}, 2018.

\bibitem{glaister2013automatic}
Jeffrey~Luc Glaister.
\newblock Automatic segmentation of skin lesions from dermatological
  photographs.
\newblock Master's thesis, University of Waterloo, 2013.

\bibitem{goyal2019scaling}
Priya Goyal, Dhruv Mahajan, Abhinav Gupta, and Ishan Misra.
\newblock Scaling and benchmarking self-supervised visual representation
  learning.
\newblock In {\em ICCV}, 2019.

\bibitem{Grandvalet2004semi}
Yves Grandvalet and Yoshua Bengio.
\newblock Semi-supervised learning by entropy minimization.
\newblock In {\em Proceedings of the 17th International Conference on Neural
  Information Processing Systems}, NIPS'04, page 529–536, Cambridge, MA, USA,
  2004. MIT Press.

\bibitem{Grathwohl2020Your}
Will Grathwohl, Kuan-Chieh Wang, Joern-Henrik Jacobsen, David Duvenaud,
  Mohammad Norouzi, and Kevin Swersky.
\newblock Your classifier is secretly an energy based model and you should
  treat it like one.
\newblock In {\em International Conference on Learning Representations}, 2020.

\bibitem{grill2020bootstrap}
Jean-Bastien Grill, Florian Strub, Florent Altché, Corentin Tallec, Pierre~H.
  Richemond, Elena Buchatskaya, Carl Doersch, Bernardo~Avila Pires,
  Zhaohan~Daniel Guo, Mohammad~Gheshlaghi Azar, Bilal Piot, Koray Kavukcuoglu,
  Rémi Munos, and Michal Valko.
\newblock Bootstrap your own latent: A new approach to self-supervised
  learning.
\newblock In {\em NeurIPS}, 2020.

\bibitem{gu2020mganprior}
Jinjin Gu, Yujun Shen, and Bolei Zhou.
\newblock Image processing using multi-code gan prior.
\newblock In {\em CVPR}, 2020.

\bibitem{he2016deep}
Kaiming He, Xiangyu Zhang, Shaoqing Ren, and Jian Sun.
\newblock Deep residual learning for image recognition.
\newblock In {\em Proceedings of the IEEE conference on computer vision and
  pattern recognition}, pages 770--778, 2016.

\bibitem{henaff2020dataefficient}
Olivier~J. Henaff, Aravind Srinivas, Jeffrey~De Fauw, Ali Razavi, Carl Doersch,
  S.~M.~Ali Eslami, and Aaron van~den Oord.
\newblock Data-efficient image recognition with contrastive predictive coding.
\newblock In {\em ICML}, 2020.

\bibitem{Hinton2007ToRS}
Geoffrey~E. Hinton.
\newblock To recognize shapes, first learn to generate images.
\newblock {\em Progress in brain research}, 165:535--47, 2007.

\bibitem{huang2020neural}
Yujia Huang, James Gornet, Sihui Dai, Zhiding Yu, Tan Nguyen, Doris~Y. Tsao,
  and Anima Anandkumar.
\newblock Neural networks with recurrent generative feedback.
\newblock In {\em NeurIPS}, 2020.

\bibitem{Huang2018weakly}
Z. {Huang}, X. {Wang}, J. {Wang}, W. {Liu}, and J. {Wang}.
\newblock Weakly-supervised semantic segmentation network with deep seeded
  region growing.
\newblock In {\em 2018 IEEE/CVF Conference on Computer Vision and Pattern
  Recognition}, pages 7014--7023, 2018.

\bibitem{huh2020transforming}
Minyoung Huh, Richard Zhang, Jun-Yan Zhu, Sylvain Paris, and Aaron Hertzmann.
\newblock Transforming and projecting images into class-conditional generative
  networks.
\newblock {\em arXiv preprint arXiv:2005.01703}, 2020.

\bibitem{hung2018adversarial}
Wei-Chih Hung, Yi-Hsuan Tsai, Yan-Ting Liou, Yen-Yu Lin, and Ming-Hsuan Yang.
\newblock Adversarial learning for semi-supervised semantic segmentation.
\newblock {\em arXiv preprint arXiv:1802.07934}, 2018.

\bibitem{isensee2018nnunet}
Fabian Isensee, Jens Petersen, Andre Klein, David Zimmerer, Paul~F. Jaeger,
  Simon Kohl, Jakob Wasserthal, Gregor Koehler, Tobias Norajitra, Sebastian
  Wirkert, and Klaus~H. Maier-Hein.
\newblock nnu-net: Self-adapting framework for u-net-based medical image
  segmentation.
\newblock {\em arXiv preprint arXiv:1809.10486}, 2018.

\bibitem{jaeger2014two}
Stefan Jaeger, Sema Candemir, Sameer Antani, Y{\`\i}-Xi{\'a}ng~J W{\'a}ng,
  Pu-Xuan Lu, and George Thoma.
\newblock Two public chest x-ray datasets for computer-aided screening of
  pulmonary diseases.
\newblock {\em Quantitative imaging in medicine and surgery}, 4(6):475, 2014.

\bibitem{ji2019invariant}
Xu Ji, João~F. Henriques, and Andrea Vedaldi.
\newblock Invariant information clustering for unsupervised image
  classification and segmentation.
\newblock In {\em ICCV}, 2019.

\bibitem{kar2019metasim}
Amlan Kar, Aayush Prakash, Ming-Yu Liu, Eric Cameracci, Justin Yuan, Matt
  Rusiniak, David Acuna, Antonio Torralba, and Sanja Fidler.
\newblock Meta-sim: Learning to generate synthetic datasets.
\newblock In {\em ICCV}, 2019.

\bibitem{karras2017progressive}
Tero Karras, Timo Aila, Samuli Laine, and Jaakko Lehtinen.
\newblock Progressive growing of gans for improved quality, stability, and
  variation.
\newblock {\em arXiv preprint arXiv:1710.10196}, 2017.

\bibitem{karras2020training}
Tero Karras, Miika Aittala, Janne Hellsten, Samuli Laine, Jaakko Lehtinen, and
  Timo Aila.
\newblock Training generative adversarial networks with limited data.
\newblock In {\em NeurIPS}, 2020.

\bibitem{karras2019style}
Tero Karras, Samuli Laine, and Timo Aila.
\newblock A style-based generator architecture for generative adversarial
  networks.
\newblock In {\em Proceedings of the IEEE Conference on Computer Vision and
  Pattern Recognition}, pages 4401--4410, 2019.

\bibitem{karras2019analyzing}
Tero Karras, Samuli Laine, Miika Aittala, Janne Hellsten, Jaakko Lehtinen, and
  Timo Aila.
\newblock Analyzing and improving the image quality of stylegan.
\newblock {\em arXiv preprint arXiv:1912.04958}, 2019.

\bibitem{kavur2020chaos}
A.~Emre Kavur, N.~Sinem Gezer, Mustafa Barış, Pierre-Henri Conze, Vladimir
  Groza, Duc~Duy Pham, Soumick Chatterjee, Philipp Ernst, Savaş Özkan, Bora
  Baydar, Dmitry Lachinov, Shuo Han, Josef Pauli, Fabian Isensee, Matthias
  Perkonigg, Rachana Sathish, Ronnie Rajan, Sinem Aslan, Debdoot Sheet,
  Gurbandurdy Dovletov, Oliver Speck, Andreas Nürnberger, Klaus~H. Maier-Hein,
  Gözde~Bozdağı Akar, Gözde Ünal, Oğuz Dicle, and M.~Alper Selver.
\newblock {CHAOS Challenge - Combined (CT-MR) Healthy Abdominal Organ
  Segmentation}.
\newblock {\em arXiv preprint arXiv:2001.06535}, 2020.

\bibitem{ke2020guided}
Zhanghan Ke, Di Qiu, Kaican Li, Qiong Yan, and Rynson W.~H. Lau.
\newblock Guided collaborative training for pixel-wise semi-supervised
  learning.
\newblock {\em arXiv preprint arXiv:2008.05258}, 2020.

\bibitem{kingma2014semi}
Diederik~P. Kingma, Danilo~J. Rezende, Shakir Mohamed, and Max Welling.
\newblock Semi-supervised learning with deep generative models.
\newblock NIPS'14, page 3581–3589, Cambridge, MA, USA, 2014. MIT Press.

\bibitem{laine2017temporal}
Samuli Laine and Timo Aila.
\newblock Temporal ensembling for semi-supervised learning.
\newblock In {\em ICLR}, 2017.

\bibitem{lee2013pseudo}
Dong-Hyun Lee.
\newblock Pseudo-label : The simple and efficient semi-supervised learning
  method for deep neural networks.
\newblock {\em ICML 2013 Workshop : Challenges in Representation Learning
  (WREPL)}, 07 2013.

\bibitem{lee2019ficklenet}
Jungbeom Lee, Eunji Kim, Sungmin Lee, Jangho Lee, and Sungroh Yoon.
\newblock Ficklenet: Weakly and semi-supervised semantic image segmentation
  using stochastic inference.
\newblock In {\em CVPR}, 2019.

\bibitem{fedsim}
Daiqing Li, Amlan Kar, Nishant Ravikumar, Alejandro~F. Frangi, and Sanja
  Fidler.
\newblock Federated simulation for medical imaging.
\newblock In {\em Medical Image Computing and Computer Assisted Intervention
  (MICCAI)}, page 159–168, 2020.

\bibitem{li2020transformation}
Xiaomeng Li, Lequan Yu, Hao Chen, Chi-Wing Fu, Lei Xing, and Pheng-Ann Heng.
\newblock Transformation-consistent self-ensembling model for semisupervised
  medical image segmentation.
\newblock {\em IEEE Transactions on Neural Networks and Learning Systems},
  2020.

\bibitem{lin2017feature}
Tsung-Yi Lin, Piotr Doll{\'a}r, Ross Girshick, Kaiming He, Bharath Hariharan,
  and Serge Belongie.
\newblock Feature pyramid networks for object detection.
\newblock In {\em Proceedings of the IEEE conference on computer vision and
  pattern recognition}, pages 2117--2125, 2017.

\bibitem{amodalVAE20}
Huan Ling, David Acuna, Karsten Kreis, Seung Kim, and Sanja Fidler.
\newblock Variational amodal object completion for interactive scene editing.
\newblock In {\em NeurIPS}, 2020.

\bibitem{lipton2017precise}
Zachary~C. Lipton and Subarna Tripathi.
\newblock Precise recovery of latent vectors from generative adversarial
  networks.
\newblock {\em arXiv preprint arXiv:1702.04782}, 2017.

\bibitem{liu2020hybrid}
Hao Liu and Pieter Abbeel.
\newblock Hybrid discriminative-generative training via contrastive learning.
\newblock {\em arXiv preprint arXiv:2007.09070}, 2020.

\bibitem{liu2018large}
Ziwei Liu, Ping Luo, Xiaogang Wang, and Xiaoou Tang.
\newblock Large-scale celebfaces attributes (celeba) dataset.

\bibitem{luc2016semantic}
Pauline Luc, Camille Couprie, Soumith Chintala, and Jakob Verbeek.
\newblock Semantic segmentation using adversarial networks.
\newblock {\em arXiv preprint arXiv:1611.08408}, 2016.

\bibitem{mendoncca2013ph}
Teresa Mendon{\c{c}}a, Pedro~M Ferreira, Jorge~S Marques, Andr{\'e}~RS Marcal,
  and Jorge Rozeira.
\newblock Ph 2-a dermoscopic image database for research and benchmarking.
\newblock In {\em 2013 35th annual international conference of the IEEE
  engineering in medicine and biology society (EMBC)}, pages 5437--5440. IEEE,
  2013.

\bibitem{Misra_2020_CVPR}
Ishan Misra and Laurens van~der Maaten.
\newblock Self-supervised learning of pretext-invariant representations.
\newblock In {\em Proceedings of the IEEE/CVF Conference on Computer Vision and
  Pattern Recognition (CVPR)}, June 2020.

\bibitem{mittal2019semi}
Sudhanshu Mittal, Maxim Tatarchenko, and Thomas Brox.
\newblock Semi-supervised semantic segmentation with high-and low-level
  consistency.
\newblock {\em IEEE Transactions on Pattern Analysis and Machine Intelligence},
  2019.

\bibitem{miyato2019vat}
T. {Miyato}, S. {Maeda}, M. {Koyama}, and S. {Ishii}.
\newblock Virtual adversarial training: A regularization method for supervised
  and semi-supervised learning.
\newblock {\em IEEE Transactions on Pattern Analysis and Machine Intelligence},
  41(8):1979--1993, 2019.

\bibitem{mondal2018fewshot}
Arnab~Kumar Mondal, Jose Dolz, and Christian Desrosiers.
\newblock Few-shot 3d multi-modal medical image segmentation using generative
  adversarial learning.
\newblock {\em arXiv preprint arXiv:1810.12241}, 2018.

\bibitem{ng2001}
Andrew~Y. Ng and Michael~I. Jordan.
\newblock On discriminative vs. generative classifiers: A comparison of
  logistic regression and naive bayes.
\newblock In {\em Proceedings of the 14th International Conference on Neural
  Information Processing Systems: Natural and Synthetic}, NIPS'01, page
  841–848, Cambridge, MA, USA, 2001. MIT Press.

\bibitem{Noroozi2016}
Mehdi Noroozi and Paolo Favaro.
\newblock Unsupervised learning of visual representations by solving jigsaw
  puzzles.
\newblock In Bastian Leibe, Jiri Matas, Nicu Sebe, and Max Welling, editors,
  {\em Computer Vision -- ECCV 2016}, pages 69--84, Cham, 2016. Springer
  International Publishing.

\bibitem{odena2016semisupervised}
Augustus Odena.
\newblock Semi-supervised learning with generative adversarial networks.
\newblock {\em arXiv preprint arXiv:1606.01583}, 2016.

\bibitem{olsson2020classmix}
Viktor Olsson, Wilhelm Tranheden, Juliano Pinto, and Lennart Svensson.
\newblock Classmix: Segmentation-based data augmentation for semi-supervised
  learning.
\newblock {\em arXiv preprint arXiv:2007.07936}, 2020.

\bibitem{park2019semantic}
T. {Park}, M. {Liu}, T. {Wang}, and J. {Zhu}.
\newblock Semantic image synthesis with spatially-adaptive normalization.
\newblock In {\em 2019 IEEE/CVF Conference on Computer Vision and Pattern
  Recognition (CVPR)}, pages 2332--2341, 2019.

\bibitem{park2020swapping}
Taesung Park, Jun-Yan Zhu, Oliver Wang, Jingwan Lu, Eli Shechtman, Alexei~A.
  Efros, and Richard Zhang.
\newblock Swapping autoencoder for deep image manipulation.
\newblock In {\em NeurIPS}, 2020.

\bibitem{perarnau2016invertible}
Guim Perarnau, Joost van~de Weijer, Bogdan Raducanu, and Jose~M. Álvarez.
\newblock Invertible conditional gans for image editing.
\newblock {\em arXiv preprint arXiv:1611.06355}, 2016.

\bibitem{pham2020meta}
Hieu Pham, Qizhe Xie, Zihang Dai, and Quoc~V. Le.
\newblock Meta pseudo labels.
\newblock {\em arXiv preprint arXiv:2003.10580}, 2020.

\bibitem{liftsplat20}
Jonah Philion and Sanja Fidler.
\newblock Lift, splat, shoot: Encoding images from arbitrary camera rigs by
  implicitly unprojecting to 3d.
\newblock In {\em ECCV}, 2020.

\bibitem{plumerault2020controlling}
Antoine Plumerault, Hervé~Le Borgne, and Céline Hudelot.
\newblock Controlling generative models with continuous factors of variations.
\newblock {\em arXiv preprint arXiv:2001.10238}, 2020.

\bibitem{Raj2019gan}
A. {Raj}, Y. {Li}, and Y. {Bresler}.
\newblock Gan-based projector for faster recovery with convergence guarantees
  in linear inverse problems.
\newblock In {\em 2019 IEEE/CVF International Conference on Computer Vision
  (ICCV)}, pages 5601--5610, 2019.

\bibitem{ranzato2011}
M. Ranzato, J. Susskind, V. Mnih, and G. Hinton.
\newblock On deep generative models with applications to recognition.
\newblock In {\em Proceedings of the 2011 IEEE Conference on Computer Vision
  and Pattern Recognition}, CVPR '11, page 2857–2864, USA, 2011. IEEE
  Computer Society.

\bibitem{richardson2020encoding}
Elad Richardson, Yuval Alaluf, Or Patashnik, Yotam Nitzan, Yaniv Azar, Stav
  Shapiro, and Daniel Cohen-Or.
\newblock Encoding in style: a stylegan encoder for image-to-image translation.
\newblock {\em arXiv preprint arXiv:2008.00951}, 2020.

\bibitem{ronneberger2015u}
Olaf Ronneberger, Philipp Fischer, and Thomas Brox.
\newblock U-net: Convolutional networks for biomedical image segmentation.
\newblock In {\em International Conference on Medical image computing and
  computer-assisted intervention}, pages 234--241. Springer, 2015.

\bibitem{rotemberg2020patientcentric}
Veronica Rotemberg, Nicholas Kurtansky, Brigid Betz-Stablein, Liam Caffery,
  Emmanouil Chousakos, Noel Codella, Marc Combalia, Stephen Dusza, Pascale
  Guitera, David Gutman, Allan Halpern, Harald Kittler, Kivanc Kose, Steve
  Langer, Konstantinos Lioprys, Josep Malvehy, Shenara Musthaq, Jabpani Nanda,
  Ofer Reiter, George Shih, Alexander Stratigos, Philipp Tschandl, Jochen
  Weber, and H.~Peter Soyer.
\newblock A patient-centric dataset of images and metadata for identifying
  melanomas using clinical context.
\newblock {\em arXiv preprint arXiv:2008.07360}, 2020.

\bibitem{sajjadi2016reg}
Mehdi Sajjadi, Mehran Javanmardi, and Tolga Tasdizen.
\newblock Regularization with stochastic transformations and perturbations for
  deep semi-supervised learning.
\newblock In {\em NeurIPS}, NIPS'16, page 1171–1179, Red Hook, NY, USA, 2016.
  Curran Associates Inc.

\bibitem{salimans2016gans}
Tim Salimans, Ian Goodfellow, Wojciech Zaremba, Vicki Cheung, Alec Radford, and
  Xi Chen.
\newblock Improved techniques for training gans.
\newblock In {\em Proceedings of the 30th International Conference on Neural
  Information Processing Systems}, NIPS'16, page 2234–2242, Red Hook, NY,
  USA, 2016. Curran Associates Inc.

\bibitem{shiraishi2000development}
Junji Shiraishi, Shigehiko Katsuragawa, Junpei Ikezoe, Tsuneo Matsumoto,
  Takeshi Kobayashi, Ken-ichi Komatsu, Mitate Matsui, Hiroshi Fujita, Yoshie
  Kodera, and Kunio Doi.
\newblock Development of a digital image database for chest radiographs with
  and without a lung nodule: receiver operating characteristic analysis of
  radiologists' detection of pulmonary nodules.
\newblock {\em American Journal of Roentgenology}, 174(1):71--74, 2000.

\bibitem{sohn2020fixmatch}
Kihyuk Sohn, David Berthelot, Chun-Liang Li, Zizhao Zhang, Nicholas Carlini,
  Ekin~D. Cubuk, Alex Kurakin, Han Zhang, and Colin Raffel.
\newblock Fixmatch: Simplifying semi-supervised learning with consistency and
  confidence.
\newblock In {\em NeurIPS}, 2020.

\bibitem{souly2017semi}
N. {Souly}, C. {Spampinato}, and M. {Shah}.
\newblock Semi supervised semantic segmentation using generative adversarial
  network.
\newblock In {\em IEEE International Conference on Computer Vision (ICCV)},
  pages 5689--5697, 2017.

\bibitem{Srivastava2012}
Nitish Srivastava and Russ~R Salakhutdinov.
\newblock Multimodal learning with deep boltzmann machines.
\newblock In F. Pereira, C.~J.~C. Burges, L. Bottou, and K.~Q. Weinberger,
  editors, {\em Advances in Neural Information Processing Systems}, volume~25,
  pages 2222--2230. Curran Associates, Inc., 2012.

\bibitem{stirenko2018chest}
Sergii Stirenko, Yuriy Kochura, Oleg Alienin, Oleksandr Rokovyi, Yuri
  Gordienko, Peng Gang, and Wei Zeng.
\newblock Chest x-ray analysis of tuberculosis by deep learning with
  segmentation and augmentation.
\newblock In {\em 2018 IEEE 38th International Conference on Electronics and
  Nanotechnology (ELNANO)}, pages 422--428. IEEE, 2018.

\bibitem{tarvainen2017mean}
Antti Tarvainen and Harri Valpola.
\newblock Mean teachers are better role models: Weight-averaged consistency
  targets improve semi-supervised deep learning results.
\newblock In {\em Advances in neural information processing systems}, pages
  1195--1204, 2017.

\bibitem{viazovetskyi2020stylegan2}
Yuri Viazovetskyi, Vladimir Ivashkin, and Evgeny Kashin.
\newblock Stylegan2 distillation for feed-forward image manipulation.
\newblock In {\em ECCV}, 2020.

\bibitem{wang2018high}
Ting-Chun Wang, Ming-Yu Liu, Jun-Yan Zhu, Andrew Tao, Jan Kautz, and Bryan
  Catanzaro.
\newblock High-resolution image synthesis and semantic manipulation with
  conditional gans.
\newblock In {\em Proceedings of the IEEE conference on computer vision and
  pattern recognition}, pages 8798--8807, 2018.

\bibitem{wang2017chestx}
Xiaosong Wang, Yifan Peng, Le Lu, Zhiyong Lu, Mohammadhadi Bagheri, and
  Ronald~M Summers.
\newblock Chestx-ray8: Hospital-scale chest x-ray database and benchmarks on
  weakly-supervised classification and localization of common thorax diseases.
\newblock In {\em Proceedings of the IEEE conference on computer vision and
  pattern recognition}, pages 2097--2106, 2017.

\bibitem{xie2020unsupervised}
Qizhe Xie, Zihang Dai, Eduard Hovy, Minh-Thang Luong, and Quoc~V. Le.
\newblock Unsupervised data augmentation for consistency training.
\newblock In {\em NeurIPS}, 2020.

\bibitem{Yeh2017}
R.~A. {Yeh}, C. {Chen}, T.~Y. {Lim}, A.~G. {Schwing}, M. {Hasegawa-Johnson},
  and M.~N. {Do}.
\newblock Semantic image inpainting with deep generative models.
\newblock In {\em 2017 IEEE Conference on Computer Vision and Pattern
  Recognition (CVPR)}, pages 6882--6890, 2017.

\bibitem{neuralmotionplanner}
Wenyuan Zeng, Wenjie Luo, Simon Suo, Abbas Sadat, Bin Yang, Sergio Casas, and
  Raquel Urtasun.
\newblock End-to-end interpretable neural motion planner.
\newblock In {\em IEEE Conf. Comput. Vis. Pattern Recog.}, June 2019.

\bibitem{Zhang2019survey}
Man Zhang, Zhou Yong, Jiaqi Zhao, Man Yiyun, Bing Liu, and Rui Yao.
\newblock A survey of semi- and weakly supervised semantic segmentation of
  images.
\newblock {\em Artificial Intelligence Review}, 53, 12 2019.

\bibitem{zhang2018unreasonable}
Richard Zhang, Phillip Isola, Alexei~A Efros, Eli Shechtman, and Oliver Wang.
\newblock The unreasonable effectiveness of deep features as a perceptual
  metric.
\newblock In {\em Proceedings of the IEEE Conference on Computer Vision and
  Pattern Recognition}, pages 586--595, 2018.

\bibitem{style3d}
Yuxuan Zhang, Wenzheng Chen, Huan Ling, Jun Gao, Yinan Zhang, Antonio Torralba,
  and Sanja Fidler.
\newblock Image gans meet differentiable rendering for inverse graphics and
  interpretable 3d neural rendering.
\newblock In {\em arXiv:2010.09125}, 2020.

\bibitem{zhang21}
Yuxuan Zhang, Huan Ling, Jun Gao, Kangxue Yin, Jean-Francois Lafleche, Adela
  Barriuso, Antonio Torralba, and Sanja Fidler.
\newblock Datasetgan: Efficient labeled data factory with minimal human effort.
\newblock In {\em CVPR}, 2021.

\bibitem{zhu2020domain}
Jiapeng Zhu, Yujun Shen, Deli Zhao, and Bolei Zhou.
\newblock In-domain gan inversion for real image editing.
\newblock {\em arXiv preprint arXiv:2004.00049}, 2020.

\bibitem{zhu2016generative}
Jun-Yan Zhu, Philipp Kr{\"a}henb{\"u}hl, Eli Shechtman, and Alexei~A Efros.
\newblock Generative visual manipulation on the natural image manifold.
\newblock In {\em European conference on computer vision}, pages 597--613.
  Springer, 2016.

\end{thebibliography}
}

\end{document}